\documentclass[twoside]{article}
\usepackage{ecj,palatino,epsfig,latexsym, natbib,apalike}
\usepackage{graphicx}
\usepackage{amssymb}
\usepackage{xcolor,colortbl}

\usepackage{xfrac}
\usepackage{slashbox}

\usepackage[english]{babel}
\usepackage{svg}
\usepackage{makecell}
\newcommand{\minus}{\scalebox{0.5}[1.0]{$-$}}

\usepackage{pgf,tikz}
\usepackage{pgfplots}
\usepgfplotslibrary{colorbrewer}
\usepgfplotslibrary{groupplots}
\usepackage{tikzscale}
\usepackage{standalone}
\usepackage{dblfloatfix}
\pgfplotsset{cycle list/Dark2-8}
\usepackage{pgfplotstable}
\pgfplotsset{compat=1.7}
\usepackage{multirow}
\usepackage{enumitem}
\usepackage{longtable}
\usepackage{placeins}

\usepackage{subfig}
\usepackage{caption}
\usepackage{longtable}

\usepackage{color}
\usepackage{algorithm}
\usepackage{color,soul}
\usepackage{multicol}
\usepackage{xcolor}
\usepackage[noend]{algpseudocode}
\usepackage{etoolbox}
\usepackage{adjustbox}% http://ctan.org/pkg/adjustbox
\usepackage{float}

\newfloat{algorithm}{!b}{lop}
\usepackage{algorithmicx}
\makeatletter
\usepackage{tabularx}

\usepackage{svg}
\usepackage{array}
\usepackage{arydshln}
\usepackage{listings}

\def\BState{\State\hskip-\ALG@thistlm}
\makeatother
\usepackage{setspace}
\usepackage{balance}
\usepackage{eqnarray}

\usepackage[acronym]{glossaries}
\usepackage{titlesec}
\titlespacing\section{0pt}{12pt plus 4pt minus 2pt}{0pt plus 2pt minus 2pt}
\titlespacing\subsection{0pt}{5pt plus 4pt minus 2pt}{0pt plus 2pt minus 2pt}
\titlespacing\subsubsection{0pt}{5pt plus 4pt minus 2pt}{0pt plus 2pt minus 2pt}

\newacronym{dt}{DT}{Decision Tree}
\newacronym{uav}{UAV}{Unmanned Aerial Vehicle}
\newacronym{ugv}{UGV}{Unmanned Ground Vehicle}
\newacronym{slam}{SLAM}{Simultaneous Localisation and Mapping}
\newacronym{ga}{GA}{Genetic Algorithm}
\newacronym{ge}{GE}{Grammatical Evolution}
\newacronym{ep}{EP}{Evolutionary Programming}
\newacronym{gp}{GP}{Genetic Programming}
\newacronym{ann}{NN}{Neural Network}
\newacronym{hh}{HH}{Hyper-Heuristics}
\newacronym{rhs}{RHS}{Right-Hand Side}
\newacronym{lhs}{LHS}{Left-Hand Side}
\newacronym{ABC}{ABC}{Artificial Bee Colony}
\newacronym{pso}{PSO}{Particle Swarm Optimisation}
\newacronym{wsn}{WSN}{Wireless Sensor Network}
\newacronym{ai}{AI}{Artificial Intelligence}
\newacronym{ailta}{AILTA}{Adaptive Iteration Limited Threshold Accepting}
\newacronym{ROS}{ROS}{Robot Operating System}
\newacronym{RTS/CTS}{RTS/CTS}{Request-to-send/ Clear-to-send}
\newacronym{fsm}{FSM}{Finite State Machine}
\newacronym{manet}{MANET}{Mobile Ad-hoc NETwork}
\newacronym{gn}{GN}{Graph Neuron}
\newacronym{dhgn}{DHGNS}{Distributed Heuristic Graph Neurons Swarm}
\newacronym{ml}{ML}{Machine Learning}
\newacronym{hrl}{HRL}{Hierarchical Reinforcement Learning}
\newacronym{slcs}{SLCS}{Swarm Learning Classifier System}
\newacronym{slcs2}{SLCS2}{Swarm Learning Classifier System 2.0}
\newacronym{lcs}{LCS}{Learning Classifier System}
\newacronym{bba}{BBA}{Bucket Brigade Algorithm}
\newacronym{TCS}{TCS}{Temporal Classifier System}
\newacronym{ZCS}{ZCS}{Zeroth-level Classifier System}
\newacronym{llh}{LLH}{Low-Level Heuristics}
\newacronym{XCS}{XCS}{eXended Classifier System}
\newacronym{SAMUEL}{SAMUEL}{Strategy Acquisition Method Using Empirical Learning}
\newacronym{tdm}{TDM}{Time-Division Multiplexing }
\newacronym{SARSA}{SARSA}{State-Action, Reward, State-Action}
\newacronym{rl}{RL}{Reinforcement Learning}
\newacronym{GRASP}{GRASP}{Greedy Randomised Adaptive Search Procedures}
\newacronym{WOLF}{WOLF}{Win Or Learn Fast}
\newacronym{DSTG}{DSTG}{Defence Science and Technoogy Group}
\newacronym{ldpl}{LDPL}{Log-Distance Path Loss}
\newacronym{snr}{SNR}{Signal to Noise Ratio}
\newacronym{snir}{SNIR}{Signal to Noise and Interference Ratio}
\newacronym{csma}{CSMA}{Carrier-Sense Multiple Access}
\newacronym{uct}{UCT}{Upper Confidence Bound applied to trees}
\newacronym{rnn}{RNN}{Recurrent Neural Network}
\newacronym{dnn}{DNN}{Deep Neural Network}
\newacronym{hgn}{HGN}{Hierarchical Graph Neurons}
\newacronym{mlp}{MLP}{Multi-Layer Perceptron}
\newacronym{TD}{TD}{Temporal Difference learning}

\newacronym{CNN}{CNN}{Convolutional Neural Network}
\newacronym{enn}{ENN}{Evolved Neural Network}
\newacronym{snn}{SNN}{Supervised training Neural Network}
\newacronym{ci}{$\mathrm{CI}_{90}$}{Confidence Interval}
\newacronym{mb}{MB}{Manually designed Behaviour}
\newacronym{tof}{ToF}{Time-of-Flight}
\newacronym{tdma}{TDMA}{Time-division Multiple Access}
\newacronym{fcc}{FCC}{Federal Communications Commission}
\newacronym{ta}{TA}{Threshold Acceptance}
\newacronym{ce}{CE}{Conformit\'{e} Europ\'{e}ene}
\newacronym{SINR}{SINR}{Signal to Interference, plus Noise Ratio}
\newacronym{des}{DES}{Discrete Event Simulation}
\newacronym{Nov}{BDMA-2}{Behaviour DoMination Algorithm 2}
\newacronym{long}{long}{long range}
\newacronym{short}{short}{short range}
\newcolumntype{L}[1]{>{\raggedright\let\newline\\\arraybackslash\hspace{0pt}}m{#1}}
\newcolumntype{R}[1]{>{\hsize=#1\hsize\raggedright\arraybackslash}X}%

\usepackage{cleveref}
\setlist{nosep}

\begin{document}

	%Data transfer via swarm robotics: a dynamic, online policy learning approach
	%	\title{Autonomous policy evolution for swarm robotics in data transfer applications}
	%\title{Behaviour creation for swarms with realistic data communication \\
	%\title{Scooby-doo and the autonomous swarm}
	\title{
		%	Cooperative Co-Evolution of swarm behaviours for enabling communication in unexplored, contested environments
		%Evolution of swarm robotic behaviours, via  sharing, for a complex task in stochastic, contested environments
		%	LCS inspired swarm behaviour creation for challenging tasks in stochastic, unexplored, contested environments
		Swarm Behaviour Evolution via Rule Sharing and Novelty Search
	}

	\ecjHeader{x}{x}{xxx-xxx}{2019}{Swarm Behaviour Evolution}{P. Smith et al.}
	\author{\name{\bf Phillip Smith} \hfill \addr{phillipsmith@monash.edu}\\
		\addr{Faculty of Information Technology, Monash University, Clayton, Victoria, Australia}
		\AND 
		\name{\bf Robert Hunjet} \hfill\addr{Robert.Hunjet@dst.defence.gov.au }\\
		\addr{Defence Science and Technology Group, Edinburgh, South Australia}
		\AND 
		\name{\bf Aldeida Aleti} \hfill\addr{aldeida.aleti@monash.edu}\\
		\addr{Faculty of Information Technology, Monash University, Clayton, Victoria, Australia}
		\AND 
		\name{\bf Asad Khan} \hfill\addr{asad.khan@sensoranalytics.com.au}\\
		\addr{Sensor Analytics, Mount Waverley, Australia}	}
	\maketitle

    \begin{abstract}

	Evolutionary Computation has proven to be successful in creating robotic agent behaviours that allow them to autonomously perform specific tasks. This area is known as evolutionary robotics and has only recently been extended to robotic swarms, where a swarm-wide behaviour emerges from individual robotic agent actions and interactions. This extension has lead to a divide in the swarm robotic community between autonomously created and designed behaviour. Evolution works have implemented simple swarm tasks in a range of conditions, and behaviour design works have implemented more challenging tasks, such as data-transfer facilitation, but within controlled conditions. In our previous work, we attempted to overcome this divide by utilising a human designed rule generator to autonomously create rule-set behaviours for this challenging data-transfer task. This algorithm saw reasonable success, however, the created behaviours were limited to unrestricted environments and the evolution search often stagnated on plateaus in the fitness landscape. Additionally, the evolution was slow, as the heterogeneous agents evolved in isolation and thus did not propagate good results to fellow swarm members.
We present in this paper an exertion of our previous work by increasing the robustness and coverage of the evolution search via hybridisation with a state-of-the-art novelty search and accelerate the individual agent behaviour searches via a novel behaviour-component sharing technique. Via these improvements, we present Swarm Learning Classifier System 2.0 (SLCS2), a behaviour evolving algorithm which is robust to complex environments, and seen to out-perform a human behaviour designer in challenging cases of the data-transfer task in a range of environmental conditions. Additionally, we examine the impact of tailoring the SLCS2 rule generator for specific environmental conditions. We find this leads to over-fitting, as might be expected, and thus conclude that for greatest environment flexibility a general rule generator should be utilised.

\end{abstract}

\begin{keywords}
	Robot Behaviour Evolution, Swarm Robotics,  Cooperative Coevolution, Reinforcement Learning, Novelty Search
\end{keywords}

\section{Introduction} \label{Intro} 
In the early 2000s, a branch of multi-agent robotics returned to the scientific community's focus, known as \textit{Swarm robotics} \citep{beni2004swarm}. The idea of this branch was to utilise a large collection of relatively simple robotic agents to perform a task together, drawing on the expression ``quantity has a quality all of its own''.
This field took inspiration from nature; looking to insects \citep{Soleymani2015} and bacteria \citep{Timmis2016} for behaviour mechanisms. Not long after this, the evolutionary computation community became interested in swarms~\citep{trianni2008evolutionary} shifting the questions from `What can we make these robots do?' to `What can these robots make themselves do?' However, over the last decade, the divide between manually designed behaviours and autonomously created behaviours has grown; specifically in terms of swarm task complexity; 
evolved swarms have been exclusively tested with trivial tasks, such as collision-free locomotion~\citep{Heinerman2015b}, while designed swarms have completed more complex tasks, such as data-transfer~\citep{fraser2017simulating}. That being said, behaviour evolution has allowed swarms to operate in various environments without further human interaction and designed behaviours are limited to controlled conditions.  As such, our research looks to bridge this gap by applying an alternative evolutionary technique for creating behaviours to solve this complex swarm task of data-transfer. This alternative approach is inspired by \gls{lcs} \citep{Holland1978, smith1980learning} and evolves behaviours consisting of high-level condition-action rules and utilising \gls{rl} during swarm operation to select from these rules. Additionally, this approach uses rules generated via a grammar-based decoder, which allows human designers to manipulate the potential of the agent behaviours without restricting flexibility. Finally, this behaviour evolution is conducted on a behaviourally heterogeneous swarm, which is a swarm of functionally identical robotic agents, each with a unique \textit{individual behaviour} controllers. This heterogeneity allows the behaviour of the overall swarm, or `\textit{emergent} behaviour', to be more diverse and flexible at solving the assigned task \cite{gomes2015cooperative}.

In our prior implementation of the above rule evolution algorithm  \citep{ smith2018data}, which we retroactively title \acrfull{slcs} 1.0 (SLCS1), we achieved reasonable success in simple environment conditions. However, this implementation saw many of the robotic evolution issues presented  \cite{silva2016open}; some evolution searches stagnated due to local optima entrapment and the behaviours of the heterogeneous swarm were slow due to isolated evolution in each agent. To overcome these issues, this paper presents \glsunset{slcs2} \glsunset{slcs} \gls{slcs2}, a significantly improved system which overcomes the stagnation in evolution by implementing a state-of-the-art novelty search, \gls{Nov} \citep{meyerson2017discovering}, in hybrid with the evolution, and by implementing a novel form of rule exchange amounts the heterogeneous swarm.

To evaluate \gls{slcs2}, behaviours are evolved for a range of challenging instances of the data-transfer task, simulated in a 2D environment. These challenges include obstacles which must be circumvented and impact signal quality, malicious networking jammers which prevents communication when agents are in range, and a realistically stochastic communication signals which makes actions non-deterministic.
This evaluation is conducted in contrast to a human-designed behaviour constructed especially for the proposed application. A human-designer is compared in this study as no prior swarm behaviour evolution has addressed such a complex task and thus cannot be directly implemented for comparisons. In addition to this main evaluation, the performance benefit of the two key additions to \gls{slcs2} are evaluated by isolating each addition and examining the fitness improvements. Finally, as the grammar-based decoder of \gls{slcs} utilises a human-designed rule generator, this paper also investigates tailoring this construction for specific problem conditions. The resulting behaviours are compared to the general behaviours in terms of swarm performance and flexibility.

In summary, the main contribution of this paper is an improved swarm behaviour evolution algorithm which allows a robotic agent swarm to complete a realistic network assistance task with performance competitive to human-designed swarm behaviours. The individual contributions of this system are:
\vspace*{-5pt}
\begin{enumerate}[noitemsep,nolistsep]
	\item the hybridisation of a behaviour evolution search and \gls{Nov} for a more robust and wider covering exploration of the behaviour-space.
	\item the development and evaluation of a novel rule sharing algorithm for a co-evolved population of individual agent behaviours. 
\end{enumerate}
In addition, we:     \vspace*{-5pt}
\begin{enumerate}[label=\alph*),noitemsep,nolistsep]
	\item develop a novel human-designed swarm behaviour for the proposed data-transfer task to validate the behaviour creation process of \gls{slcs2} in terms of swarm effectiveness and flexibility.
	\item evaluate the performance and flexibility impacts of specialising the rules of \gls{slcs2} for specific problems within the data-transfer domain.
\end{enumerate}

The remainder of this paper is organised as follows: \Cref{litRev} presents related works on behaviour evolution; \Cref{approach} presents the architecture of the utilised swarm agents, including environment observations, available actions, and \gls{rl}. \Cref{evoApproach} presents the proposed evolution algorithm, including the grammar-based rule structure, the hybridisation with \gls{Nov}, and the rule sharing technique during agent behaviour evolution.
\Cref{exp} presents the simulation experiments used for validating \gls{slcs2}, along with the design and validation of a human-made behaviour for comparison to \gls{slcs2}, and three grammar alterations to specialise the evolution for specific problem conditions.
In  \Cref{RandD} the experiment results of this study are presented and discussed, and finally \Cref{conc} conclude this study, makes final remarks on the results and presents intended further work.

\section{Related Work}\label{litRev}
In this review, we first outline existing swarm behaviour evolution, highlighting the limitations in flexibility or resulting complexity.  As presented in \Cref{Intro}, these approaches are limited to simplistic behaviours, and thus, this review further explores non-swarm behaviour evolution and behaviour decoding methods to achieve the desired complexity. In addition, this review explores methods for extending swarm evolution via behaviour exchange and state-of-the-art behaviour-space searches.

\subsection{Behaviour Evolution}    \label{lib_evo}
One of the earliest works that investigate the evolution of swarm behaviours is the work by \cite{perez2003}. A simple swarm behaviour was created via a \gls{ga}, which utilised three manually designed action-condition rules. As these rules were specially designed for the given problem the swarm could solve non-trivial tasks. However, only one environment was explored and only 9 behaviours were possible, which shows this work did not explore task flexibility in the swarm. 

In contrast to this high level design, \cite{trianni2008evolutionary,nelson2004,hauert2009} and \cite{Heinerman2015b} evolved \gls{ann} behaviours which linked sensory inputs to motor controls. These low-level or \textit{primitive} \citep{duarte2018evolution} controllers allowed for large variation in behaviour creation, which permitted environment and task flexibility. However, such large behaviour-spaces came at the cost of limiting the swarm task to simple requirements, such as aggregation \cite{trianni2008evolutionary} or locomotion \cite{Heinerman2015b}. This issue has been noted by \cite{duarte2018evolution} which stated such controllers are insufficient for `complex robots in tasks beyond mere locomotion' and is observed in most robotic evolution works \citep{silva2016open}. Additionally, this reduced potential is seen in the work by \cite{hauert2009}, which explored \gls{ann} swarm behaviours for data transfer. Due to the lack of swarm complexity, Heuert limited the swarm to a pre-defined behaviour and evolved only a controller for the agents' turning magnitude to fit this behaviour.

To overcome such complex behaviour spaces, another form of evolution, \gls{lcs} \citep{Holland1978, smith1980learning}, has been used in single-agent behaviours in \cite{grefenstette1988,grefenstette1995, Gordon1995, Hurst2002}; and \cite{smith2005rcs}. These behaviours often consisted of non-primitive condition-action rules which the agents consulted at each decision point. This consultation consists of disregarding rules with untrue conditions and selecting from the remaining actions. Online \gls{rl} was implemented for this final selection process which reduces the responsibility of the evolution, as it allowed the evolved rule-set to contain multiple action responses to each condition, which the \gls{rl} was responsible for selecting from. This is unlike \gls{ann} which needed to account for every condition and evolve the behaviour controller to respond appropriately.  Furthermore, unlike the rule evolution of Perez, these robotic \glspl{lcs} evolved or dynamically generated each rule, allowing reasonable flexibility in terms of environment and problem conditions addressed. By evolving and learning these rules in a heterogeneous swarm, \gls{lcs} is seen to be similar to the Tangled Program Graphs of \cite{kelly2018emergent}, however, an \gls{lcs} swarm would explore each behaviour in parallel.

In relation to the tasks these \gls{lcs} robotic controllers are created for, many are closely related to operations seen in swarm robotics. These include:
environment navigation \citep{grefenstette1988}, enemy evasion \citep{Gordon1995}, target tracking \citep{ grefenstette1995}, and photo-taxis \citep{Hurst2002} in single-agent or small-scale multi-agent operations. However, no works, to our knowledge, have explicitly explored swarm robotic based \gls{lcs}. \label{robotLCS} 

From this review of evolution algorithms, it is found previous swarm behaviour evolution works either lacked flexibility in the behaviour creation range, lacked complexity in the behaviours created, or over-relied on the evolution to create a behaviour controller which accounts for all agent experiences.  To overcome these issues, \gls{lcs} has been identified as an effective robotic behaviour creation and adjustment algorithm. Therefore, this study explores \gls{lcs} for swarms.

\subsection{Rule Decoding}
As part of implementing rule-based behaviours, the decoding of a random digit string genomes to conditions and actions must be explored. 
In     \cite{Hurst2002} a transitional \gls{lcs} approach was taken which saw the genome directly represent environment patterns via sensor readings.
In contrast, \cite{Gordon1995} implemented a form of \gls{gp} which indirectly decoded conditions with non-primitive attribute limits via a specificity hierarchy with an ``\textit{attribute} within \textit{range}'' structure.

As an extension to \gls{gp}, a decoding process by \cite{ONeill2003} known as \textit{Grammatical generation} shows potential for rule generation. Grammatical generation showed greater flexibility in generation by utilising a human-defined grammar structure to determine how each digit of the genome was decoded based on the prior digits' decoding. Using this approach allowed any form of controller to be implemented.

\subsection{Behaviour Exchange}
In addition to the evolution technique above, a method for the swarm to cooperatively evolve heterogeneous behaviours is examined in this work. In most swarm evolution studies, this process was not investigated as behaviourally homogeneous swarms were utilised. That is, all agents were equipped with identical behaviours. These behaviours underwent crossover and mutation but were evaluated in isolation. In contrast, \cite{Heinerman2015b} evolved a population of behaviours simultaneously across a swarm of robotic agents via both individual and \textit{social} learning. Heinerman had swarm members which achieved a fitness above a defined threshold broadcast their behaviour genome to surrounding neighbours. Upon receipt of this genome, agents trialled the behaviour.  Such a process proved to have significant fitness improvement across the swarm. However, the swarm application of this work did not require cooperative behaviours, and thus this genome interchange did not explore behaviour specialisation or cooperation \cite{gomes2015cooperative}. Additionally, this work did not explore partial behaviour exchange between agents.

In contrast to this swarm-based behaviour exchange, a non-robotic multi-agent \gls{lcs} evolution was conducted in \cite{takadama2000} which implemented rule exchange between cooperative agents.     This approach limited the exchange to only an elite subset of the agents' rule-set and upon receipt of this subset, agents would replace the lowest scoring rules of their own behaviour. This exchange partially moved the recipient rule-set toward the globally optimal rule-set, allowing exploration along the behaviour-space vector. However, this exchange was only conducted between random agent pairs, rather than the elitist exchange of \cite{Heinerman2015b}. As such, high-performing members could be lost during this exchange process.

From this review, it is shown limited exploration in rule exchange has been made for swarm behaviours. However, a study in non-robotic multi-agent \gls{lcs}  may be utilised. Some alterations in this \gls{lcs} rule exchange are required to prevent high-performance and specialised agents being lost.

\subsection{Behaviour-space Search}
As a third and final area of examination in swarm behaviour evolution, behaviour-space search algorithms are reviewed. 

In the above evolution works, fitness-based searches were implemented which accepted or rejected new behaviours and \gls{lcs} rules via the resulting task performance. Such approaches allowed high-performing solution neighbourhoods to be focused during the search but risked local optima entrapment.  This fitness-based approach was also taken in our prior work, \gls{slcs}1 \citep{ smith2018data}, which utilised \gls{ailta} \citep{Miser2010}. \gls{ailta} allowed \gls{slcs}1 to often find effective behaviour solutions, however, challenging environments would often see the evolution stagnate in local optima for many generations.

In contrast to these fitness greedy searches, novelty-based searches and novelty-fitness hybrid searches were explored for swarm behaviours in \cite{gomes2013evolution}. This study found novelty-based and fitness-based had situational domination over one another, but a novelty-fitness hybrid outperformed both. This novelty-fitness hybrid implemented a simple sum equation to derive a behaviour's score during evolution. 

As an extension to novelty-fitness hybrids, \cite{meyerson2017discovering}, \gls{Nov}, proposed a multi-objective search with a fitness-novelty Pareto front for single-agent behaviour search. This algorithm allowed faster exploration of the behaviour-space than exclusive fitness searches and higher behaviour fitness discovery than exclusive novelty search. Additionally, this algorithm outperformed four other fitness-novelty multi-objective search algorithms.    \gls{Nov} achieved this improved behaviour-space exploration by comparing two solutions, \textit{x} and $y$, via the domination effect function,
\begin{equation}\vspace*{-4pt}
e(x,y)= f(x)-f(y)-\omega \cdot \Delta b(x,y) \label{domination}
\end{equation} 
where \textit{f(x)} was the fitness of solution \textit{x}, \textit{w} was a weight for equalising the fitness and novelty scale, \textit{b(x)} was the behaviour-space location of \textit{x}, and thus $\Delta b(x,y)$ was the euclidean distance between the two solutions.  Solution \textit{x} was said to dominate \textit{y}, or `$x>y$', if $e(x,y)>0$. In contrast, if $e(x,y) \le 0$ than \textit{x} did not dominate \textit{y}, or `$x\ngtr y$', though this was not equivalent to $y>x$.  
Using this domination principle, \gls{Nov} controlled the intake of a solution, \textit{x}, into a collection, \textit{Y}, and the removal of dominated solutions in \textit{Y} via    
\begin{eqnarray}
\nexists y \in Y, e(y,x) >0 \Rightarrow  Y\cup \{x\} \to Y  \label{nov_iclude}\\ 
\forall y \in Y, \iff e(x,y)>0 \Rightarrow Y\backslash \{y\} \to Y \label{nov_remove}  
\end{eqnarray}
Using this domination approach, the search may explore any number of solutions along the Pareto front. A similar implementation of novelty-fitness hybrid search was implemented in robotic swarms by \cite{gomes2013evolution}. However, this study utilised  a novelty-fitness score sum for $(\mu+\lambda)$ evolution rather than the \gls{Nov} algorithm.

From this review, it is seen that most swarm evolution work has been limited to fitness-based search. However, novelty-fitness searches have been seen to discover greater solutions in fewer iterations and the novelty-fitness Pareto front algorithm \gls{Nov} has seen considerable success for single agents. As such, the inclusion of \gls{Nov} in the proposed system is expected to both improve search performance and be a novel contribution to the swarm behaviour evolution field. However, such inclusion requires the function $b(x)$ be defined for the proposed rule-set behaviours and the weighting value \textit{w} be tailored for the problem and rule-sets.

\section{Background: Agent design}\label{approach}

Before the evolution process of \gls{slcs2} is presented, the agent design is discussed for context. This discussion is broken into three key areas: agent knowledge, action control, and \gls{rl}. These agent attributes are discussed within the context of the data-transfer problem domain and thus we briefly define the task and terminology.

Swarm agents are tasked with  transferring \textit{packets}  from a \textit{source} to a \textit{sink}, both of which are \textit{network-nodes}. These nodes may be human-operated machines or other autonomous agents.  The task is simulated in a time-discrete simulation and thus observations and actions are conducted within \textit{time-steps}.  The term \textit{packet} is used to abstractly refer to a data-segment which can be transferred within a time-step and swarm agents are limited to one \textit{packet} being held at a time. The \textit{source} is a networking device which has data to send and the \textit{sink} is the destination of this data. In addition, the environments have \textit{obstacles} and \textit{jammers}. Obstacles are objects which the agents must avoid collision with. Signals can be sent through these obstacles, though will be reduced in strength. Jammers are malicious networking devices which detect communication and produce noise on the signal channel. They are included as an expected issue for the swarm to overcome. 

The agents of this study each hold a set of \textit{rules}. Each rule, $\gamma$, features a \textit{condition} and \textit{action}, and an agent's rule-set, $\Gamma$, dictate the possible behaviour of the agent.

\subsection{Agent Knowledge}
As discussed in \Cref{Intro}, each member of a swarm is relatively simple in terms of both sensor and actuator complexity. In this study, simplicity is achieved by agent knowledge being limited to internal information, surrounding neighbour information, $N$, and memory of previously seen or shared neighbours, $K$. This knowledge is collected via a lidar and a communication device. It can be noted that this study attempts to keep the communication protocol as abstract as possible to allow future real-world implementations flexible in such selection. Each time-step, agents update $N$, detect if a packet is held in the communication buffer, and calculate wireless interference strength, $NI$. Additionally, agents periodically update $K$ via knowledge propagation. A detailed discussion of the agent knowledge gathering is given in Appendix A.1.

\subsection{Agent Actions}
Agents of this study have two types of action: movement and communication. These actions respectively have the agent reposition itself in the environment relative to other swarm members or network-nodes in $K$, or attempt a packet transfer with a neighbour in $N$. Further detail of these actions is also given in Appendix A.2.

\subsection{Reinforcement Learning}
To determine which action to execute each time-step, the rule-set of agent \textit{i}, $\Gamma_{i}$, utilises Q-learning \cite{watkins1989learning} to predict each rule's quality. As such, each rule, $\gamma_{n,i}$, holds a  quality value, $q_{n,i}$, in addition to the condition, $c_{n,i}$, and  action, $a_{n,i}$. To being \gls{lcs} rule selection, a short-list of valid rules, $\Gamma'_{i} \subset \Gamma_{i}$, is made such that $\forall \gamma{n,i} \in \Gamma'_i : \tilde{S_i} \implies c_{n,i}$. Following this, a rule is selected based on $q_{n,i}$. In this study, \gls{GRASP} \citep{resende2016optimization} is utilised for rule selection. This process is defined as, 
\begin{equation}
\begin{split}
q_{\mathrm{grasp}} =   q_{\max,i}-\alpha_{\mathrm{grasp}}(q_{\max,i}-q_{\min,i})\\
\Gamma''_i \subset \Gamma_i', \ \gamma_{n,i} \in \Gamma''_i:  \ q_{n,i} \ge q_{\mathrm{grasp}} \\
p(\gamma_{n,i})=
\begin{cases}
\frac{1}{|\Gamma''_i|} & \gamma_{n,i}\in \Gamma''_i\\
0 & \mathrm{otherwise} \\
\end{cases}
\end{split}
\label{graspEQ}
\end{equation}
where $q_{\max,i}$ and $q_{\min,i}$ are respectively the maximum and minimum $q$ of $\Gamma'_{i}$, $\alpha_{\mathrm{grasp}}$ is the exploration-exploitation ratio and $p(\gamma_{n,i})$ is the probability of rule $n$ being selected.  

After the selected rule's action is performed a reward is calculated via geographical routing \citep{Ghafoor2014}. This reward, $\rho$, is defined as,

\begin{equation}
\begin{split}
\rho_p = & \sum_{p \in \mathrm{Pa}_h} \Delta(d_{\mathrm{sink},p}) + \sum_{p \in \mathrm{Pa}_t} \Delta(d_{\mathrm{sink},p}) +\sum_{p \in \mathrm{Pa}_r} \Delta(d_{\mathrm{sink},p})\\
\rho_s =& \begin{cases}
0, & |\mathrm{Pa}_h|> 0  \vee |\mathrm{Pa}_t|> 0 \vee |\mathrm{Pa}_r|> 0\\
\Delta(d_{\mathrm{source},i}), & \mathrm{otherwise}
\end{cases}\\
\rho =& \log(|\rho_p| + 1)\cdot \mathrm{sgn}(\rho_p) + \rho_s - C_a \label{reward}
\end{split} \hspace{-10pt}
\end{equation}
where $\mathrm{Pa}_h$, $\mathrm{Pa}_t$ and $\mathrm{Pa}_r$ are the packets held, transferred and received in the time-step;  $\Delta(d_{\mathrm{sink},p})$ is the change in distance between the packet and sink since last scoring; $\Delta(d_{\mathrm{source},i})$ is equivalent for source and agent; and $C_a$ is a cost for the performed action type (movement or communication). The three packet collections encourage the agents to transport packets toward the sink, send packets to neighbours, and positions themselves receive packets from neighbours. By rewarding packet-less agents for moving towards the source, agent do not become trapped in local-minima by $\rho_p=0$. % Without this reward $C_a$ must be tailored for each environment size. 
By setting a cost for each action type, the swarm is discouraged from learning behaviours which are energy expensive.

Using the calculated reward, Q-learning updates the estimated quality of the rule selected in the time-step, $\gamma_{t,i}$, and all rules in $\Gamma'_i$ with matching actions. That is
\begin{equation}
\begin{split}
\gamma_{n,i} \in \Gamma''_{i}: a_{n,i} = a_{t,i}\\ 
\forall \gamma_{n,i} \in \Gamma''_{t,i},    q_{n,i} \cdot (1-\alpha_q)+\alpha_q (\rho+ \beta_q \cdot \max( q_{t+1,i}) ) \to q_{n,i}\end{split}
\end{equation}
where $\alpha_q$ is the learning rate, $\beta_q$ is the future discount factor and $\max(q_{t+1,i})$ is the maximum $q$ of $\Gamma'_{i}$ in the following time-step.

\section{Behaviour Evolution}\label{evoApproach}

\begin{figure}
	\centering
	\includegraphics[width=\linewidth]{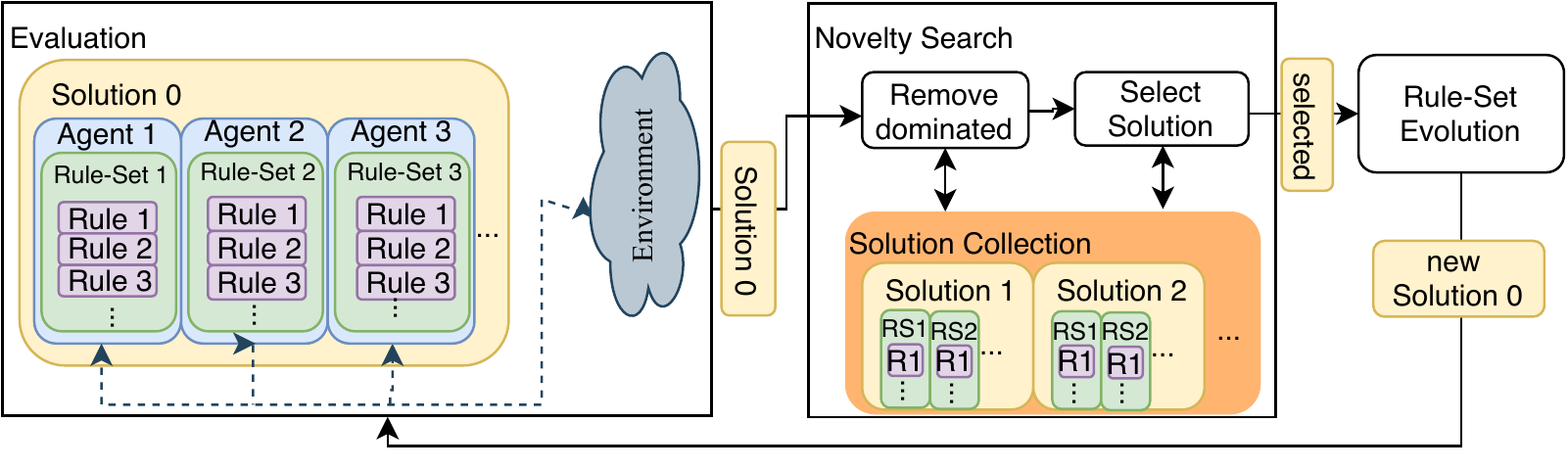}\vspace*{-2mm}
	\caption{Evolution overview. Each solution is evaluated by equipping the swarm agents with the respective rule-set and observing the resulting swarm interaction with the environment. After evaluation, the solution, the resulting swarm fitness, and the agent contributions are sent to the novelty search. This process determines if the new solution should be held for further exploration and if any existing solutions should be removed from the collection. One solution of this collection is then nominated for further exploration. That solution is altered by the rule-sets being evolved and the resulting new solution is sent for evaluation, completing the cycle. }
	\label{fig:evolution-overview}
\end{figure}

Having established the foundation of the swarm operation in the previous section, this section can now present the main contribution of this paper, the evolution algorithm for creating an effective set of agent rule-sets for a heterogeneous swarm. This evolution search is continued until a terminating criterion is reached. Each cycle consists of an evaluation, a solution-collection update via novelty domination,  and a new \textit{solution} being created via the evolution of an existing solution's rule-sets. This process is depicted in \Cref{fig:evolution-overview} which also presents our terminology for the three layers of controller hierarchy. That is, a \textit{solution} is an ordered set of \textit{rule-sets}, where each \textit{rule-set} belongs to an agent of the swarm. These \textit{rule-sets} then contain \textit{rules}, and these \textit{rules} contain \textit{conditions} and \textit{actions} to dictate what an agent should do when environmental or internal states are observed. In this study, these rules are generated via random digit string being decoded via a genetic grammar \citep{ONeill2003}.

The proposed algorithm utilises both \gls{Nov} \citep{meyerson2017discovering} and rule/behaviour exchange \citep{Heinerman2015b, takadama2000}, however both are seen as novel iterations. 
For the implementation of \gls{Nov}, the novelty equation must be defined as no work has implemented this algorithm in swarms, or on rule-based behaviours.
For the exchange process, this work uses a fitness proportional exchange with partial rule-set importation via crossover. Prior studies have implemented complete behaviour importation \citep{Heinerman2015b} or exchanged between random neighbours \citep{takadama2000}.

The remainder of this section is broken into defining the grammar-based rule generation process, and the three components of the behaviour evolution search: evaluation, solution-collection update and rule-set evolution.

\subsection{Rule Generation}
To create the initial solution for exploration, each agent is given a rule-set; each rule-set consists of $|\Gamma|$ rules. This rule-set size was defined as 100 (as listed in \Cref{Values}), which was empirically found to give flexibility in rule selection each time-step whilst not overly slowing the RL convergence. This relation was found from empirical investigation of $|\Gamma'|$ each time-step, finding it to scale in complexity approximately $O(|\Gamma|^{w}), w:(0,1)$ and $|\Gamma'|$ to linearly relate to the available action (for flexibility) and sub-linearly relate to convergence time. 

Additionally, new rules are generated in the rule-set evolution to create new solutions. These rules are generated via the grammar in \Cref{mainGrammar} which utilised the observations and actions presented in Appendix A and defines $P_rs$ and $d_th$ as respectively a signal noise and distance threshold. The content of this grammar allows the rules to be high-level,  with neither conditions or actions utilising primitive controls. 

\begin{table}
	\small      
	\centering
	\caption{Basic Rule Grammar. $<>$ denotes non-terminals which require further decoding. Each decoding option is separated by $|$.} \label{mainGrammar}    \vspace*{-2mm}
	\begin{tabularx}{\linewidth}{|@{}L{0.26\linewidth}@{}lR{1}|}
		\hline
		\textless Condition\textgreater & $\rightarrow$ & \textless Condition\textgreater \& \textless Condition\textgreater\ \mbox{\textbar\ NI $\le$ $P_{\mathrm{rs}}$} \mbox{\textbar \ \textless Packet\textgreater} \ \mbox{\textbar \ \textless Neigh. Type\textgreater} \mbox{\textbar \ \textless Distance\textgreater}  \\
		\textless     Packet \textgreater      &$\rightarrow $&  $\exists p_i$  \ \textbar \  $\nexists p_i$  \\
		\textless Neigh. Type\textgreater& $\rightarrow$ & $\exists n \in N: \mathrm{type}_{n,i} =$ network-node \ \mbox{\textbar $\nexists n \in N: \mathrm{type}_{n,i} =$network-node} \  \\
		\textless Distance\textgreater & $\rightarrow$ &\textless RHS\textgreater  \textless Operator\textgreater  \textless LHS\textgreater \\
		\textless RHS \textgreater  & $\rightarrow$ &$d_{\mathrm{so},i}$  \ \textbar \  $d_{\mathrm{si},i}$  \ \mbox{\textbar \ $ d_{c,i}$}  \ \mbox{\textbar \  $d_{\mathrm{sink},i}$ } \ \mbox{\textbar \ $ d_{\mathrm{source},i} $}\ %%\mbox{\textbar \  $d_{\upsilon_,i} $} \ \mbox{\textbar \  $d_{o,i}$}
		\\
		\textless Operator\textgreater  & $\rightarrow$ & $<$  \ \textbar \ $>$ \ \textbar \ =  \\ 
		\textless LHS \textgreater  &$ \rightarrow$ &$d_{\mathrm{so},i}$  \ \textbar \  $d_{\mathrm{si},i}$  \ \textbar \ $ d_{c,i}$  \ \textbar \  $d_{\mathrm{th}}$ \ \mbox{\textbar\ $\mathrm{net}$} \ 
		\\ \hline
		\textless Action\textgreater & $\rightarrow$ & Move \textless Direction\textgreater \textless Move target\textgreater   \ \textbar \ \textless Networking \textgreater \\
		\textless Direction\textgreater & $\rightarrow$ & Toward \ \textbar \  Away   \ \textbar \  Orbit \\
		\textless Move target\textgreater & $\rightarrow$ & so\ \textbar\ si\ \textbar\ c  \ \textbar \  source\ \textbar\  sink\ \textbar \  $\upsilon_{\mathrm{close}}$  \mbox{\textbar\  $o$ } \\
		\textless Networking\textgreater  & $\rightarrow$ & \textless Collect\textgreater   \ \textbar \   \textless Send\textgreater\\
		\textless Collect\textgreater & $\rightarrow$ & Collect from source \\
		\textless Send\textgreater &  $\rightarrow$ & Send to\textless Send target\textgreater \\
		\textless Send target\textgreater & $\rightarrow$ & so \ \textbar\ si\ \textbar\ c  \ \textbar \  source   \ \textbar \  sink  \\
		\hline 
	\end{tabularx}
\end{table} 
\label{grammar}

\gls{slcs} has the condition and action of each rule stored as a sequence of digits. Decoding each rule from digit string to usable function starts in the form `\mbox{\textless Condition\textgreater $\implies$ \textless Action\textgreater}', where each `$<>$' pair denotes a \textit{non-terminal} and all other words, phrases or symbols are \textit{terminals}.  The digits are sequentially used to convert the leftmost non-terminal, using the grammar of \Cref{mainGrammar}, until only terminals remain. To decode a non-terminal \Cref{mainGrammar} presents conversion options, separated by `$|$', with the digit acting as a wrapping index via the modulus of the options available. 

For a more detailed discussion of this grammar decoding, the reader is encouraged to view O'Neill's original paper \citep{ONeill2003}. Additionally, an example of rule generation with this grammar is shown in Appendix B. It is noted this grammar structure is only used for rule-generation, and evolution techniques are not conducted on the rules themselves in this study (they are conducted on the rule-set).

\subsection{Evaluation}
To evaluate a solution, a new instance of the swarm operation is performed with the rule-sets of the solution utilised by the respective swarm agents. In this study, the agents of \Cref{approach} are implemented, and thus a new instance has all agents start with no $K$ knowledge and all rule \textit{q} values equal to 0. The agents use \gls{rl} to optimise the provided rule-sets for solving the task until a terminating criterion for the operation is met. At termination, evaluation is conducted at both the solution and rule-set level. Additionally, the individual rules of each agent are assessed throughout the operation. These three evaluations are presented below and utilised in the remainder of this section.

\subsubsection{Solution Evaluation}
To evaluate the quality of a solution, the swarm performance in the assigned task is measured via a single fitness metric. In this study, this swarm fitness is, 
\begin{equation}
\mathrm{fit}_{\boldsymbol{\Gamma}} = \frac{p_s}{p} - \frac{T_s}{T}, \mathrm{fit}_{\boldsymbol{\Gamma}} \in \mathbb{R}:(-1,1) \label{fitness_equations}     
\end{equation}
where $p_s$ and $T_s$ are the packets that reached the sink and the time-steps completed in the operation, respectively. $\mathrm{fit}_{\boldsymbol{\Gamma}}>0$ indicates the swarm can solve the task within the time-limit, and as $\mathrm{fit}_{\boldsymbol{\Gamma}} \rightarrow 1$ the swarm is seen to be highly effective. $\mathrm{fit}_{\boldsymbol{\Gamma}}<0$ indicates the swarm has failed the task, though $\mathrm{fit}_{\boldsymbol{\Gamma},<0} \rightarrow 0$  suggests data transfer is being accomplishing, but at an unacceptably slow rate. 

\subsubsection{Rule-set Evaluation}
In addition to the overall solution evaluation, each rule-set is assessed by measuring the corresponding agent's contribution to the task. This evaluation allows the evolution process to differentiate the rule-sets which are leading the swarm to success, and the rule-sets which have the respective agent hinder the operation. 
\label{rule_setEval}

For each agent of this study, contribution is defined as the number of packets, $\tilde{p_i}$, which the agent interacted with during the operation (held for at least one time-step) and reached the sink by operation termination, and $\bar{\tilde{p}}$ is the mean contribution  across the swarm. Using this metric, the agents, and thus the rule-sets, are broken into two groups; high-quality, \textit{hq}, and low-quality, \textit{lq}, 
\begin{eqnarray}
\forall \Gamma_i \in \boldsymbol{\Gamma}_{hq}, \tilde{p_i} \ge \bar{{\tilde{p}}} \\
\forall \Gamma_i  \in \boldsymbol{\Gamma}_{lq},  \tilde{p_i} < \bar{{\tilde{p}}} \label{rule-setEval}
\end{eqnarray}    

\subsubsection{Rule Evaluation}
At the lowest level of evaluation, each rule is assessed in relation to overall strength, $\zeta$, and predicted reward error, $\epsilon$. These additional rule attributes are updated during the operation, along with the \gls{rl}. 

For $\zeta$, a limited back-propagation learning is utilised similar to TD($\lambda$) \citep{Szepesvari2010}, however, back-propagation is limited to $\mathrm{BP}_{\mathrm{max}}$ past rules. Such limitation reduces the action memory of agents and produces equivalent results, provided $\lambda^{\mathrm{BP}_{\mathrm{max}}} \rightarrow 0$. During operation, $\zeta$ is updated via,
\begin{equation}
\forall \gamma_{n,i} \in \Gamma''_{i,t-t'}, \forall t' \in \{0\dots \mathrm{BP}_{\mathrm{max}} \},  \zeta_{n,i} +\lambda^{t'} \cdot \log_{10}(|\rho|+1) \cdot \mathrm{sgn}(\rho) \to \zeta_{n,i}  
\end{equation}
where \textit{t} is the current time-step, $t'$ is the back-propagation depth and thus $\Gamma''_{t-t'}$ is $\Gamma''$ from  $t'$ time-steps in the past. 
At operation end $\zeta$ undergoes a final update via,
\begin{equation}
\forall \gamma_{n,i} \in \Gamma_{i}, \frac{\zeta_{n,i}}{\tau{n,i}} \to \zeta_{n,i}
\end{equation}
where $\tau_{n,i}$ is a usage count of rule \textit{n} during the operation.

The accuracy of each rule is measured via the error between expected reward, \textit{q}, and received reward, $\rho$, as seen in \gls{XCS} \citep{wilson1978division}. This error value is updated during operation via,
\begin{equation}
\forall \gamma_{n,i} \in \Gamma''_{i}, \epsilon_{n,i}\cdot (1-\alpha_q) + \alpha_q |\rho-q_{n,i}| \to \epsilon_{n,i}
\end{equation}
where $a_q$ is the learning rated used for the Q-learning.

These two rule attributes allow Lamarckian principles to be employed when evolving the rule-sets.

\subsubsection{Re-Evaluation}
To improve the accuracy of the above measurements, including reducing the impact of the stochastic environments, re-evaluations are conducted on solutions reporting high performance. That is, any solution which achieves $\mathrm{fit}_{\boldsymbol{\Gamma}} > 0$ is re-tested in the swarm operation. For such solutions, all evaluation measurements are taken as the mean of the recordings from each swarm implementation. This re-deployment process is repeated until the mean $\mathrm{fit}_{\boldsymbol{\Gamma}}$ falls below 0 or a limit of $\kappa$ re-deployments is reached. This extended evaluation comes at the cost of longer search execution, potentially increasing the run-time of \gls{slcs2} by  $O(\kappa)$. However, it also prevents the search exploring area of low performing solution due to a swarm reporting a statistically unlikely high result.

\subsection{Solution-Collection update}
As shown in \Cref{fig:evolution-overview}, after each evaluation of a new solution, solution 0, it is passed to the Novelty Search controller which is an implementation of \gls{Nov} and controls the content of a collection of solutions. This collection contains solutions for further exploration.  The controller determines if any domination conflicts are raised by the new solution and deleted the dominated. This implementation of the novel search also controls which solution in the collection should be explored by the evolution next. 

As presented in \Cref{litRev}, \gls{Nov} allows a search to explore a solution-space via both novelty and fitness. In this study, the inclusion of \gls{Nov} allows the evolution search to determine if a solution 0 should be added to the held solution collection, and thus undergo further exploration, via (\ref{nov_iclude}), and if prior found solutions should be removed from the collection due to being dominated by solution 0, as defined in (\ref{nov_remove}). In this study the domination equation, (\ref{domination}), is defined via $f$ being the fitness in (\ref{fitness_equations}) and $\Delta b(\Gamma_X,\Gamma_Y)$ being,
\begin{equation}	\vspace*{-1mm}
\begin{split}
\Delta b(\Gamma_X,\Gamma_Y) &= \frac{1}{|\boldsymbol{\Gamma}|} \sum_{n=0}^{n=|\boldsymbol{\Gamma} |} \Delta b'(\Gamma_{X,n},\Gamma_{Y,n}),\\ |\boldsymbol{\Gamma}|&=|\boldsymbol{\Gamma}_X|=|\boldsymbol{\Gamma}_Y|   
\\[0.3cm]
\Delta b'(\Gamma_{X,n},\Gamma_{Y,n}) &=   \frac{ \sum_{\gamma_{n} \in \Gamma_x}    b''(\gamma_{n},\Gamma_y)+\sum_{\gamma_{n} \in \Gamma_y}
	b''(\gamma_{n},\Gamma_x)  }{\lvert \Gamma_x \rvert+ \lvert \Gamma_y \rvert}     
\\[0.5cm]
b''(\gamma_\alpha,\Gamma_\beta) &=
\begin{cases} 
0, & \exists  \gamma_{n} \in \Gamma_\beta,\ (c_{\alpha} = c_{n} \wedge a_{\alpha} = a_{n}) \\[0.3cm]
1, & \mathrm{otherwise}
\end{cases}
\end{split}   \vspace*{-1mm} 
\end{equation}

This novelty comparison is conducted over all three layers of the \gls{slcs2} control hierarchy, $\Delta b$ is conducted on the solution layer, $\Delta b'$ is a comparison between agent rule-sets, and $b''(\gamma_\alpha,\Gamma_\beta)$ defines if rule $\gamma_{\alpha}$ is present in rule-set $\Gamma_\beta$. 

It can be noted, this comparison is conducted on a per-rule-set basis. That is, \textit{$\Gamma_1$} of \textit{solution X} is only compared to  \textit{$\Gamma_1$} of \textit{solution Y}. Such a limit was implemented as inter-rule-set comparisons (comparing \textit{$\Gamma_1$ } to $\Gamma_2 \dots \Gamma_{|\boldsymbol{\Gamma} |}$) was found to bias the search toward single-agent rule-sets undergoing large mutation steps, rather than multiple rule-sets undergoing smaller, controlled steps. Additionally, such a comparison encouraged behavioural heterogeneity for the sake of novelty, as opposed to the intended goal of heterogeneity for greater behavioural coverage and task flexibility.

As the final definition for implementing \gls{Nov} in this swarm evolution, this study finds $\omega$ of (\ref{domination}) effectively controls the intake and removal of solutions by being inversely proportional to the collection size. This relation encourages the intake of new solutions when the collection is small, which allows faster escape from local-optima, and increase the range of removal when the collection is large, preventing ineffective exploration.  

To utilise the resulting collection of fitness-novelty dominating solutions, \gls{slcs2} stores an exploration counter, $\varphi$,  for each solution. After the \gls{Nov} collection update, a copy of the solution which has undergone fewest evolution attempts is passed to the evolution algorithm. This selection encourages equal exploration of all stored solutions. \label{Noval}

\begin{figure}
	\centering
	\includegraphics[width=0.7\linewidth]{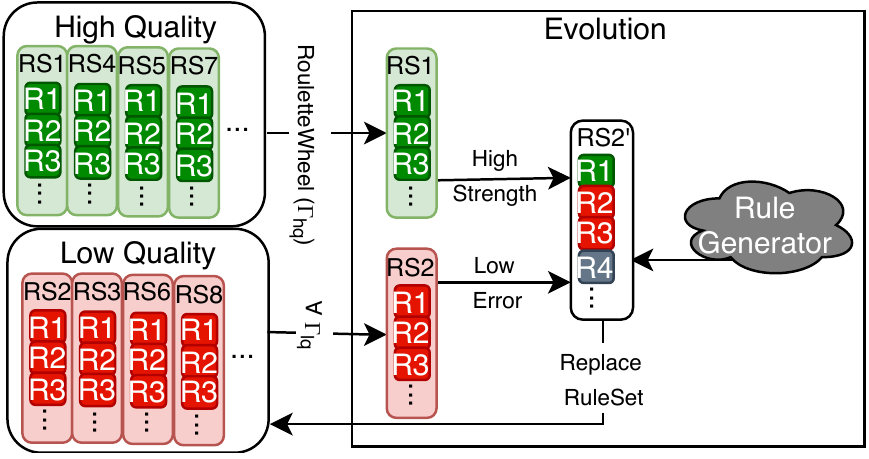}\vspace*{-2mm}
	\caption{Evolution via rule exchange and random rule generation. Example shows rule-set of Agent 2 undergoing Lamarckian evolution via removal of high error rules, inclusion of high strength rules of Agent 1, and random rules being created. This new rule-set then replaces the rule-set of Agent 2 in the solution.}
	\label{fig:cross}
\end{figure}

\subsection{Rule-Set Evolution}

To create a new solution, neighbouring a known optima, the held solution is passed to the \gls{slcs} evolution algorithm; rules are copied from high-quality \mbox{rule-sets} to low-quality \mbox{rule-sets}, as defined in (\ref{rule-setEval}). The low-quality set includes these rules, and newly generated rules, via elitist replacement. This approach is equivalent to using two-parent selection in standard evolution and using Lamarckian influenced crossover and mutation.

Lamarckian evolution has been proven invalid in biology but found to be effective within computer-science \citep{grefenstette1995,castillo2006lamarckian}. In contrast to Darwinian evolution, which relies on random change, Lamarckian evolution encourages the transfer and survival of attributes and behaviour traits heavily utilised or found effective during a solutions life-time. As such the evolving agent uses the rule evaluation to guide the crossover and mutation. 

\subsubsection{Rule Exchange / Parent Selection}
As shown in \Cref{fig:cross}, all rule-sets of low-quality agents are selected for evolution. This is equivalent to the first crossover parent being selected via an inverse-elitist selection. 

For each of these evolutions, a high-quality agent is selected proportional to $\tilde{p_i}$ (as defined in \Cref{rule_setEval}) to transfers a copy of their rule-set to the low-quality agent for crossover. This transfer is equivalent to the second parent being selected via roulette wheel.

\subsubsection{Crossover}
The Lamarckian-style crossover combines the existing low-quality rule-set with the received high-quality rule-set by creating an offspring with selective rules based on low prediction error and high strength. This process is depicted in \Cref{fig:cross}.

From the low-quality rule-set, $\varpi$ rules with the lowest $\epsilon$ (error) values are transferred to a single offspring. $\varpi$ is $|\Gamma| \cdot  \max(1,  \sfrac{\mathrm{fit}_{\boldsymbol{\Gamma}}}{2} -\sfrac{\varphi}{\varrho}), \varpi \in \mathbf{Z} \ge 0 $,
where $\varphi$ is the exploration count of the solution (see \Cref{Noval}) and  $\varrho$ is a constant weight.

From the high-quality received rule-set,  $\varsigma$ rules with the highest $\zeta$ (strength) values are transferred to the offspring. $\varsigma$ is $(|\Gamma_i| - \varpi )\cdot \mathrm{rand}(0,1), \varsigma \in \mathbf{Z} \ge 0$,
and $\mathrm{rand}(0,1)$ is a uniform random value between 0 and 1. This random value leaves space in the rule-set for newly generated rules to be added.  

The survival of low-error rules is demonstrated to be effective in \cite{Wilson1995} as the rule-set evolves to contain rules with close condition-action associations. That is, rules which hold some subject commonality between condition and action. The importation of high-strength rule is demonstrated to be effective in \cite{Holland1978}, as the rule-set evolves to contain rules which effectively contribute to solving the task. 
As an example of these two attributes, consider the rule ``\textit{if} distance from neighbour in sink direction is less than estimated network range \textit{then} send a packet to that neighbour'' ($ d_{si,i} < net \implies \mathrm{Send}(so)$).  Such a rule may receive high $\rho$ in most cases, and thus possess a high strength, but the agent does not need to hold a packet for this rule to be true. As such, this send action will have cases of failure and the rule will have a high error-rate. In contrast, the rule ``\textit{if} packet held \textit{then} request packet collection from source'' ($\exists p_i \implies \mathrm{Collect(source)}$) will always receive a low reward and thus have a low strength, but the consistency of this outcome will lead to a low error-rate and can guide the agent against making such an action when a packet is held. 

This combination of strength and error selection was chosen after empirical trials showed it to be superior in this application to exclusive strength selection, exclusive error selection and error selection from high-quality/strength selection from low-quality.  

\subsubsection{Mutation}
To achieve the effect of standard evolution mutation, the remaining $|\Gamma_i|-\varsigma-\varpi$ rules of the offspring rule-set are generated via random digit strings being produced and the grammar of \Cref{grammar} utilised to create rules. 

\subsubsection{Rule Population control}
After each rule-set offspring is created via the above crossover and mutation, it replaces the rule-set of the low-quality agent in the solution. This population control ensures a one-to-one relation between rule-sets and agents in the swarm.

\section{Experiment Setup}\label{exp}

To validate the improvement of \gls{slcs2}, and confirm if behaviours can be evolved for a non-trivial task, data-transfer, experiments are conducted in a discrete-time simulator,  Mason \citep{mason}, with 2D environments which include jammers and obstacles.  In these simulations the impact of including \gls{Nov} and rule exchange are examined, the evolved behaviours of \gls{slcs2} and a custom human-designed behaviours are compared in relation to swarm fitness and flexibility, and finally, the impact of altering the rule grammar for specific problems is examined. In this section the environments which the experiments are conducted in are described, the swarm behaviour comparison process is presented, and the three experiments of this study are defined.

\subsection{Test Environment}
The environments of this study all feature a single, stationary source device with a set number of packets, $p$, and a stationary sink device. In this study, the swarm is deployed surrounding the source and provided an estimated position of the sink. This information allows the swarm to focus on the data-transfer task, rather than solve a target search sub-task during each operation. Two source-sink distances are tested in this study, \textit{short-range} and \textit{long-range}.  The former has the source and sink devices located $600m$ apart, while the latter has them 1km apart.  In short-range tests the time-limit of an implementation, $T$ in (\ref{fitness_equations}), is 20k steps and in long-range $T$ is 40k.

For each of the above network-node ranges, four environment types are implemented:
\textbf{open}, an obstacle-free (except other agents) environment;
\textbf{jammed}, an environment with a stationary jammer randomly placed;
\textbf{urban}, an environment with randomly generated buildings;
\textbf{urban}, jammed the latter two environments combined

\begin{figure}
	\centering   
	\begin{minipage}[c]{0.5\linewidth}   
		\subfloat[]{
			\centering
			\includegraphics[width=0.3\linewidth]{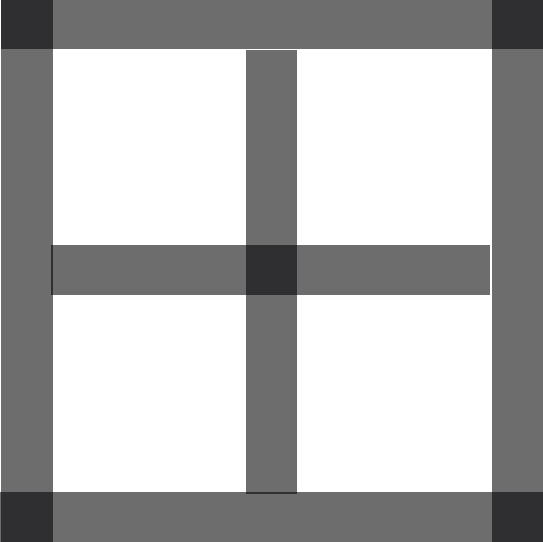}
		}
	\end{minipage}    
	\begin{minipage}[c]{0.4\linewidth}   
		\subfloat[]{\centering
			\includegraphics[width=0.7\linewidth]{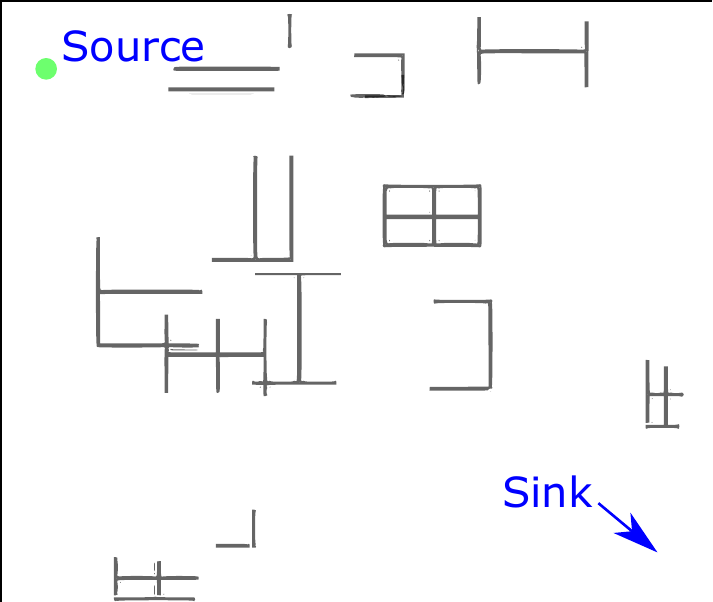}
		}
	\end{minipage}    \vspace*{-2mm}
	\caption{(a) Potential placement of the six walls to form each building. b) Example $100\times100m$ section of urban environment. Section shows 13 buildings with 1 to 6 walls.}
	\label{fig:environment}
\end{figure}

In jammed environments, a single jammer is randomly placed in the environment. This positioning is restricted such that neither the sink nor source are affected. To represent a small, concealable device, the jammer $P_t$ is equal to the swam agents'.  
Each time-step the jammer scans all communication channels and produces noise on the first detected to be active. As the standard swarm implementation has no control of the communication channel, this results in all surrounding agents failing to receive signals due to low \gls{snir}.

In urban environments a random set of buildings, in the range $\mathbb{Z}:(1,\frac{d_{\mathrm{source},\mathrm{sink}}}{10})$ is generated, each with length and width randomly in the range $\mathbb{R}:(10,30)$m. Each building is made of 6 possible walls, each in one of the positions shown in \ref{fig:environment} \textit{(a)}, each wall is 1m thick and included in the building with probability $p(0.5)$. An example environment section with these buildings is shown in \Cref{fig:environment} \textit{(b)}.  %Additionally, the building deployments are not limited by  overlap restrictions, as depicted in an environment example in Figure \ref{fig:environment} \textit{(b)}. As such, multiple buildings may connect to form large, complex structures for the swarm to circumvent and avoid entrapment.

Finally, the experiments of this study use the variable settings presented in Appendix C for knowledge sharing, motor control, \gls{rl} and evolution. Additionally, this table defines the \gls{ldpl} parameters of this study. 

\subsection{Experiment process}

For each experiment of this study, the compared algorithms are implemented in multiple environment instances for each type-range pair. Each instance has a unique random seed for agent initial positions, rule generation, network stochasticity,  jammer location (if applicable) and obstacle locations (if applicable). For the evolution process, a limit of 500 generations is imposed. The optimal fitness of each instance is recorded and the mean and 90\% \gls{ci} is reported for each algorithm in the eight type-range pairs. Additionally, the mean fitness over all environment types is reported for each algorithm. This overall fitness, demonstrating the flexibility of the algorithms to create behaviours for various conditions.

Additionally, to validate algorithm difference, the Mann-Whitney U-test is conducted for each experiment, with the p-values for each algorithm pair presented. Using these results, along with the mean and \gls{ci}, this study defines one algorithm to be \textit{significantly superior} to another iff a higher mean is reported, no \gls{ci} overlap is observed and the p-value of the pair is less than $0.1$. If either of the latter two conditions is not met we define the higher mean algorithm as \textit{non-significantly superior}.

\subsection{\gls{Nov} and Rule exchange Contributions}
To validate the contribution of \gls{Nov} and rule exchange during evolution, three swarm evolutions are examined in this first experiment:  \gls{slcs2} with \gls{Nov} replaced by the \gls{ailta} search of \gls{slcs}1, \gls{slcs2} without rule exchange during evolution, and the full \gls{slcs2}.
This experiment is conducted in 10 environment instances for each environment range and type combination.

\subsection{\gls{slcs2}, Human-designed Comparisons}
For the main evaluation of \gls{slcs2}, a human-designer is tasked with creating a behaviour for the proposed environments and the behaviours evolved by \gls{slcs2} are compared. This comparison examines the fitness difference in each environment range and type with 30 instances explored, and examines the overall mean fitness of the two approaches to asses the swarm flexibility.  Additionally, this comparison presents key findings on swarm position convergence and agent motion to justify observed differences in fitness. This human/evolution comparison is similar to the U-human tests of AutoMoDe \citep{francesca2015}, however, no time limit is placed on the human, the resulting behaviour must be operational in multiple environments.  

To improve comparability, the designed behaviour is developed as a sequence of  \textit{if-then-else} rules using the observations and actions available to the agents. However, the swarm is behaviourally homogeneous, as heterogeneous design is $O(n^\Phi)$ complex which is an unrealistic task for a human operator. Additionally, to emulate this behaviour being designed in controlled conditions and then deployed for operation without re-design, the human creation process is limited to testing the behaviour in three environment instances per environment range-type pair. The resulting behaviour design and a brief validation of this behaviour's performance compared to prior designed behaviours is presented in Appendix D.

The resulting mean and standard deviation performance of this behaviour during the design process is shown in \Cref{alphaTesting}. These results show the behaviour is operational in all test conditions.

\begin{table}[t]\centering
	\caption{Mean and SD of fitness results during design testing. Behaviour shown to effectively operate in all type and range conditions.} \label{alphaTesting}\vspace*{-2mm}
	\begin{tabular}{|l|c|c|c|c|}
		\hline \multirow{2}{*}{ \backslashbox{\textbf{Type}}{\textbf{Range}}} &\multicolumn{2}{c|}{ \textbf{Short}} &\multicolumn{2}{c|}{ \textbf{Long}} \\ \cline{2-5}
		&  \textbf{Mean} & \textbf{SD} &\textbf{Mean} & \textbf{SD}  \\ \hline
		\textbf{Open}    &    0.565 & 0.001   &    0.611 &0.004    \\\hline
		\textbf{Jammed}    &    0.641 &0.050     &    0.716 & 0.035 \\\hline
		\textbf{Urban}    &    0.555 & 0.017    &    0.573    &0.045 \\\hline
		\textbf{Urb. Jam.}    &    0.515 & 0.082    &    0.684   &0.001 \\ \hline
	\end{tabular}
\vspace*{-2.8mm}	
\end{table}

\subsection{Grammar specialisation}

For the third experiment of this study, three alterations are made to the behaviour grammar to improve performance in specific conditions. The first two alterations incorporate solutions to reduce network interference and congestion. The third alteration combines the prior two. These three alterations are presented in Tables \ref{TableChannel}-\ref{grammar1.3} as extensions of \Cref{mainGrammar}. By extending the grammar, the potential behaviours of the swarm are extended for specific problems. However, the behaviour-space is larger, which requires more exploration by the evolution search.

\begin{table}
	\small	
	\caption{Grammar: Channel, allowing transmissions to be sent on three network frequencies.}\label{TableChannel}\vspace*{-2mm}
	\begin{tabularx}{\linewidth}{|L{0.26\linewidth}@{}lR{1}|}
		\hline 
		\textless  Noise \textgreater &$\rightarrow $&   Noise on     \textless Channel\textgreater $\le$ $P_{rs}$ \\
		
		\textless LHS \textgreater  &$ \rightarrow$ &$d_{\mathrm{so},i}$  \ \textbar \  $d_{si,i}$  \ \textbar \ $ d_{c,i}$  \ \textbar \  $d_{\mathrm{th}}$ \ \mbox{\textbar\ $\mathrm{net}$} \ 
		\mbox{\textbar \ $\xi_{\upsilon_i}$}  \\
		\textless Collect\textgreater & $\rightarrow$ & Collect from source, \textless Channel\textgreater \\
		\textless Send\textgreater & $\rightarrow$ & \textless Send target\textgreater \textless Channel\textgreater \\
		\textless Channel\textgreater & $\rightarrow$ & 1   \ \textbar \  6   \ \textbar \  11 \\ \hline
	\end{tabularx}  
	\vspace*{2mm}
	\caption{Grammar: Power, allowing transmissions to be at lower power} \label{TablePower}\vspace*{-2mm}
	\begin{tabularx}{\linewidth}{|L{0.26\linewidth}@{}lR{1}|}
		\hline
		\textless LHS \textgreater  &$ \rightarrow$ &$d_{\mathrm{so},i}$  \ \textbar \  $d_{\mathrm{si},i}$  \ \textbar \ $ d_{c,i}$  \ \textbar \  $d_{\mathrm{th}}$ \ \mbox{\textbar\ $\mathrm{net}$} \ 
		\mbox{\textbar \ $\xi_{\upsilon_{i}}$} \\
		\textless Collect\textgreater & $\rightarrow$ &  Collect from source, \textless Power\textgreater \\
		\textless Send\textgreater & $\rightarrow$ & \textless Send target\textgreater \textless Power\textgreater \\
		\textless Power\textgreater  & $\rightarrow$ & 50\%   \ \textbar \  100\% \\ \hline
	\end{tabularx}\vspace*{2mm}
	\caption{Grammar: Both, combination of Channel and Power control} \label{grammar1.3}\vspace*{-2mm}
	\begin{tabularx}{\linewidth}{|L{0.26\linewidth}@{}l R{1}|}
		\hline 
		\textless Noise \textgreater &$\rightarrow $&   Noise on \textless Channel\textgreater $\le$ $P_{\mathrm{rs}}$ \\
		
		\textless LHS \textgreater  &$ \rightarrow$ &$d_{\mathrm{so},i}$  \ \textbar \  $d_{\mathrm{si},i}$  \ \textbar \ $ d_{c,i}$  \ \textbar \  $d_{\mathrm{th}}$ \ \mbox{\textbar\ $net$} \ 
		\mbox{\textbar \ $\xi_{\upsilon_{i}}$} \\
		\textless Collect\textgreater & $\rightarrow$ & Collect from source, \textless Channel\textgreater\ \textless Power\textgreater \\
		\textless Send\textgreater & $\rightarrow$ & \textless Send target\textgreater \textless Channel\textgreater\ \textless Power\textgreater \\
		\textless Channel\textgreater & $\rightarrow$ & 1   \ \textbar \  6   \ \textbar \  11 \\
		\textless Power\textgreater  & $\rightarrow$ & 50\%   \ \textbar \  100\% \\ \hline
	\end{tabularx}
\vspace*{-2mm}
\end{table}

The first alternation, Grammar:Channel, extends the agents to control which wireless frequency is transmitted on each time-step. This channel control, and required receiver synchronisation, assumes agents are equipped with a network adaptor for each channel. This extension aims to assist the swarm to overcome jamming devices and is based on the works of \cite{medal2016} which represent networking devices and jammers as an `attacker-defender game' across both physical space and the communication frequency spectrum.  The second alternative, Grammar:Power, allows agents to send signals at full or half power. This control assumes agents have access to adaptor TX settings. This grammar is based on the networking strategy of \cite{ramanathan2000topology} for overcoming inter-network signal congestion when devices are densely positioned in an environment.

Each of these grammar alternatives is implemented in 30 environment instances for each environment type and range pair. The results are compared to the main grammar in relation to swarm performance and environment flexibility.

\section{Results and Discussion}
\label{RandD}
\subsection{\gls{Nov} and Rule exchange Contributions}
\begin{figure}[b]
	\centering
	\includegraphics[height=1.8in]{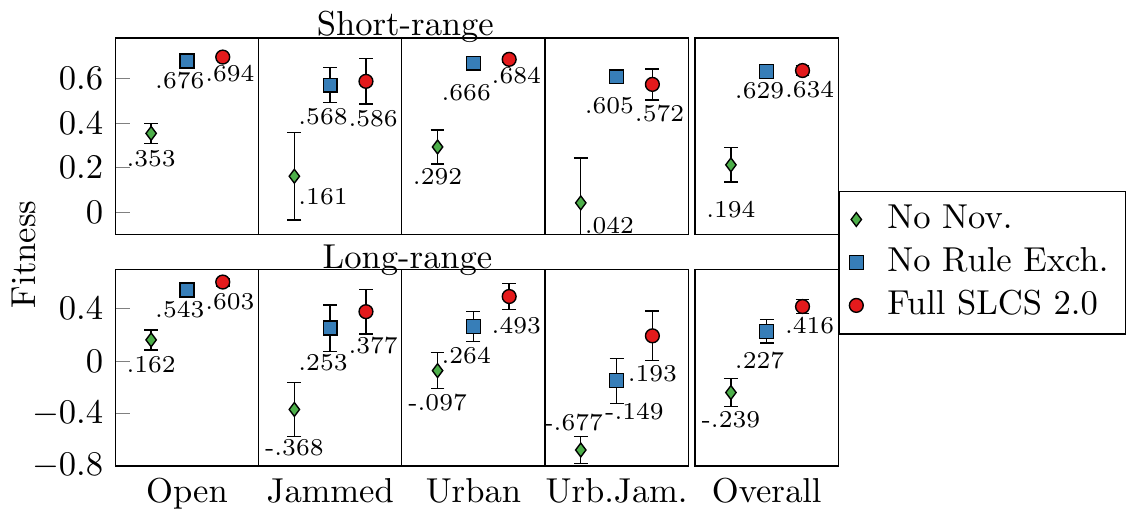}\vspace*{-2mm}
	\caption{Fitness comparison of \gls{slcs2} without novelty search, without rule exchange and full \gls{slcs2}. Removal of novelty search has large fitness reduction. Removal of rule exchange has minor reduction which is statistically significant in some cases. }
	\label{fig:noswarpnonov}
\end{figure}

\Cref{fig:noswarpnonov} presents the initial validation of \gls{slcs2} via comparison without novelty search and rule exchange. These results show that novelty search makes a considerable contribution to the solution exploration and social evolution rule exchange provides a small improvement which is statistically significant in some cases.

For the novelty search, a reduction of up to 43\% of the fitness range is seen in its absence (long-range urban jammed, $\frac{\minus 0.677-0.193}{1-  \minus 1}$). Additionally, all environment settings have a Mann-Whitney p-value below 0.1 (see Appendinx E) and non-overlapping \gls{ci}, making all results statistically significant.

The removal of rule exchange during evolution, making all behaviour evolution isolated, sees a reduction in all environment setting mean fitnesses of between 0.8 and 17\% of the fitness range. However, some of these reductions are relatively small and non-statistically significant. From the Mann-Whitney results (Appendinx E) and examining the \gls{ci} overlap, it is seen statistically different results are produced in long-range open and urban settings, and thus in overall long-range. By only these cases seeing significant improvement, social evolution is shown to primarily contribute in more challenging tasks. The simpler, short-range, cases can have behaviour solutions evolved without an extensive exploration of the behaviour-space. Additionally, these results suggest environments with jammers limit the potential of social learning. 

From these initial evaluations, it is shown the primary extensions to the evolution of \gls{slcs2} have allowed for swarm behaviours with overall greater task performance and thus both novelty search and rule exchange are contributing additions.

\subsection{\gls{slcs2}, Human-designed  Comparison}\label{man_evo_results}
\begin{figure}
	\centering
	\includestandalone[mode=image, height=1.8in]{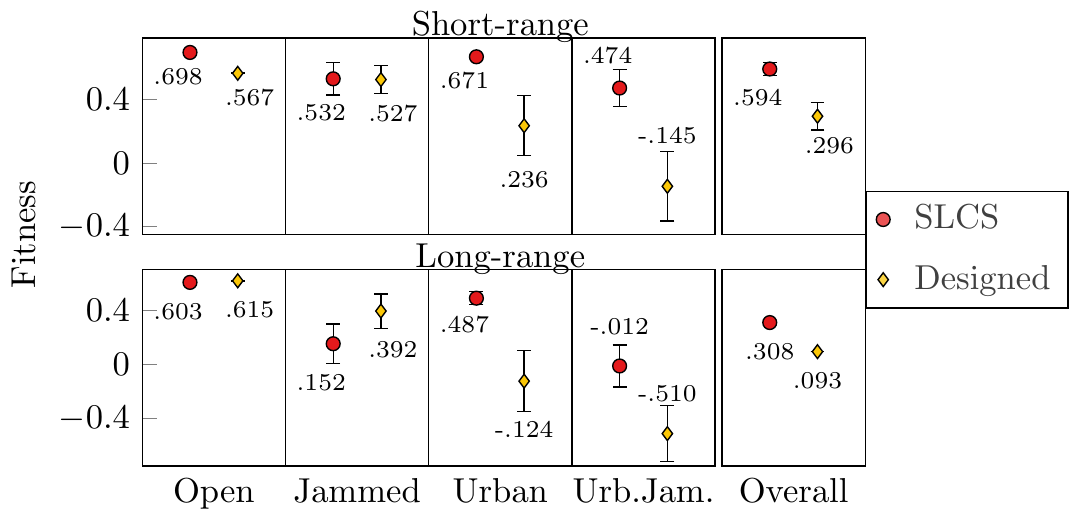}
	\vspace*{-2mm}
	\caption{Graphical results of mean fitness and \gls{ci} for \gls{slcs2} and designed behaviour. Top row is short-range,  long-range bottom.  \gls{slcs2} fitness is significantly superior in the four Urban cases, and the Overall results. However, noticeable reduction in fitness is seen for \gls{slcs2} in Jammed cases.}\label{main_results}
	\vspace*{-5mm}
\end{figure}

\Cref{main_results} show the mean fitness results and \gls{ci} for \gls{slcs2} evolved and human-designed behaviours. Additionally, Mann-Whitney results are in Appendinx E. In open environments, the short-range cases have \gls{slcs2} significantly superior to the designed behaviour, while in long-range the designed behaviour has higher mean fitness. However, in this latter case, the designed behaviour is non-significantly superior (\gls{ci} overlaps).
In jammed environments, \gls{slcs2} behaviours have a noticeable reduction in mean fitness compared to open environments. In the short-range case, this reduction makes \gls{slcs2} results similar to the designed swarm. However, in the long-range jammed experiments, the \gls{slcs2} swarm is seen to have a  reduction of 22.6\% of the fitness range, compared to long-range open \gls{slcs2}. The design behaviour is therefore noticeably (though non-significantly) superior in such situations. 
%This relatively equal comparison is also seen in short Jammed environments, with the jammer impairing the abilities of the long-range evolved swarms by 75\% (compared to the open environment), more than the designed swarm reduction of 36\%. However, the jammer caused fitness instability in both cases; leading to confidence overlap.
Finally, in the four environment settings with obstacles, the designed behaviour has large fitness reductions compared to the open setting. In contrast, \gls{slcs2} swarms have relatively similar fitness means between open and urban settings. This results in significantly superior fitness by \gls{slcs2}, with \gls{slcs2} producing higher fitnesses by 21\% to 31\% of the full fitness range.  %The evolved fitness reduction in urban jammed can be seen to be due to the jamming, 

From these results, it is demonstrated that \gls{slcs2} produces higher-performing swarm behaviours in the majority of environments compared to a human-designed behaviour and thus has wider environment flexibility. These results are confirmed by the \textit{overall} fitness, which show \gls{slcs2} evolved behaviours to be significantly superior to the designer's solution. We see a mean fitness improvement of 14.9\% ($\frac{0.594-0.296}{1 - {\minus 1}}$) in short-range and 10.7\% in long-range. The remainder of this subsection further explores the large fitness differences in short-range open environments, environments with a jammer present, and urban environments. 

\begin{figure}
	\centering
	\subfloat[]{\includegraphics[width=0.25\linewidth]{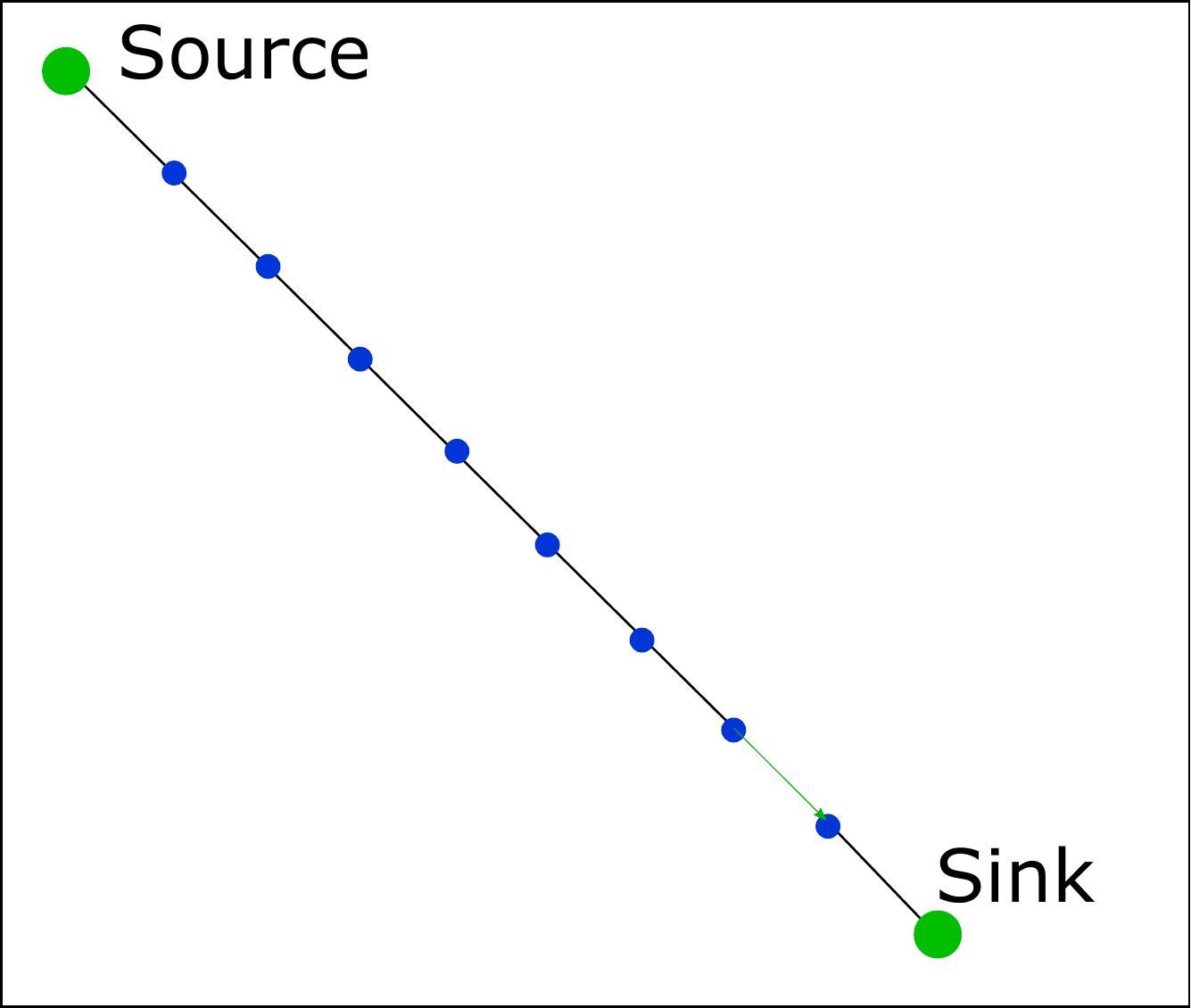} }
	\hspace*{2mm}
	\subfloat[]{    \includegraphics[width=0.25\linewidth]{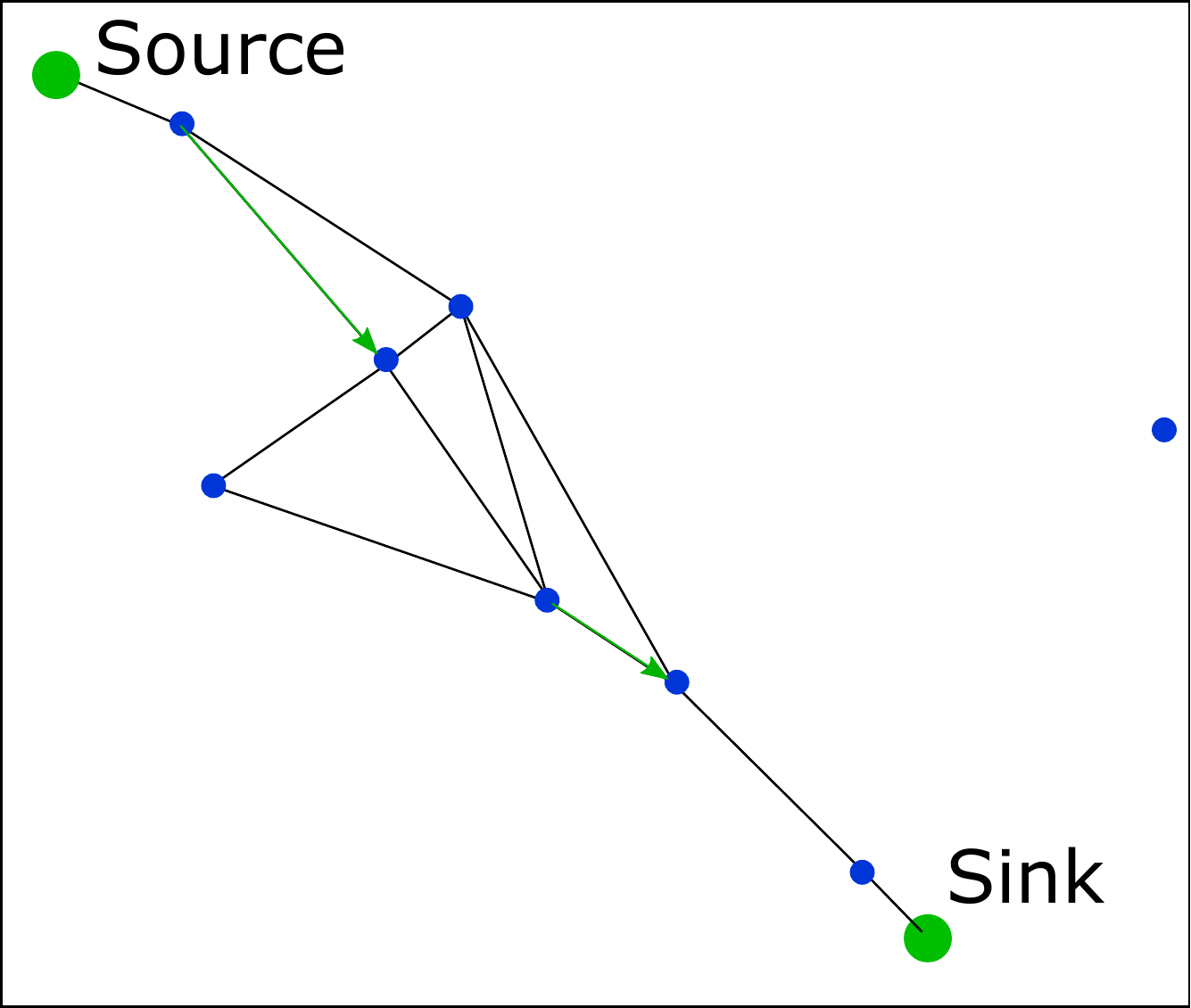} }\vspace*{-2mm}
	\caption{Short-range swarm layout in matching operation. a) Designed and b) \gls{slcs2}. Source and Sink denoted by green circles, swarm agents by blue circles, neighbour connections denoted by black lines, and data transfers in the captured time-step denoted by green arrows. Designed agents are overly close and interfere with one another. \gls{slcs2} agents spread out, contrary to intuitive design, to avoid internal interference.   } \label{short_range_comp}
\end{figure}

Further exploring the success of \gls{slcs2} in open, short-range environments, \Cref{short_range_comp} shows an example of the designed swarm and an \gls{slcs2}-evolved swarm after converging to stationary locations in the environment. As shown in \textit{a)}, the designed behaviour has agents evenly spaced between source and sink. This proximity causes agents to create network interference for one-another, reducing the transmission potential. In contrast, the \gls{slcs2}-evolved swarm, \textit{b)}, spread-out perpendicular to the optimal packet path. This increases the distance between agents, thus reducing intra-swarm interference. This novel behaviour shows the creativity of the \gls{slcs2} evolution allows for behaviours beyond the ideas of the human designer. Furthermore, in \textit{b)}, one agent in the \gls{slcs2} swarm moves out of the connection chain, and remains idle. This further reduces the intra-swarm interference at the cost of the agent's contribution to the transfer and thus individual performance. Such an outcome demonstrates the altruistic potential of evolved heterogeneous swarm members.

%    \begin{figure}
%        
%        \centering
%        \includegraphics[width=0.8\linewidth]{figures/nojammrestrition}
%        %\subfloat[]{    \includegraphics[height=1.2in]{figures/jammrestrition}}
%        \caption{Count of rules being valid ($c_{n} \implies \tilde{S}$)  for random agent during operation. Left, operation with jammer. Agent interacts with jammer at time-step ~500. Valid rules are reduced to a small subset, which is repeatedly explored for remainder of operation. No action in subset allows for agent escape. Right, operation with no jammer. Majority of rules have reasonably high validity over operation. Non-linear plots indicate set of valid rules changes as observations change. }
%        \label{fig:jammrestrition}
%    \end{figure}

\begin{figure}    
	\centering
	\subfloat[]{\includegraphics[width=0.25\linewidth]{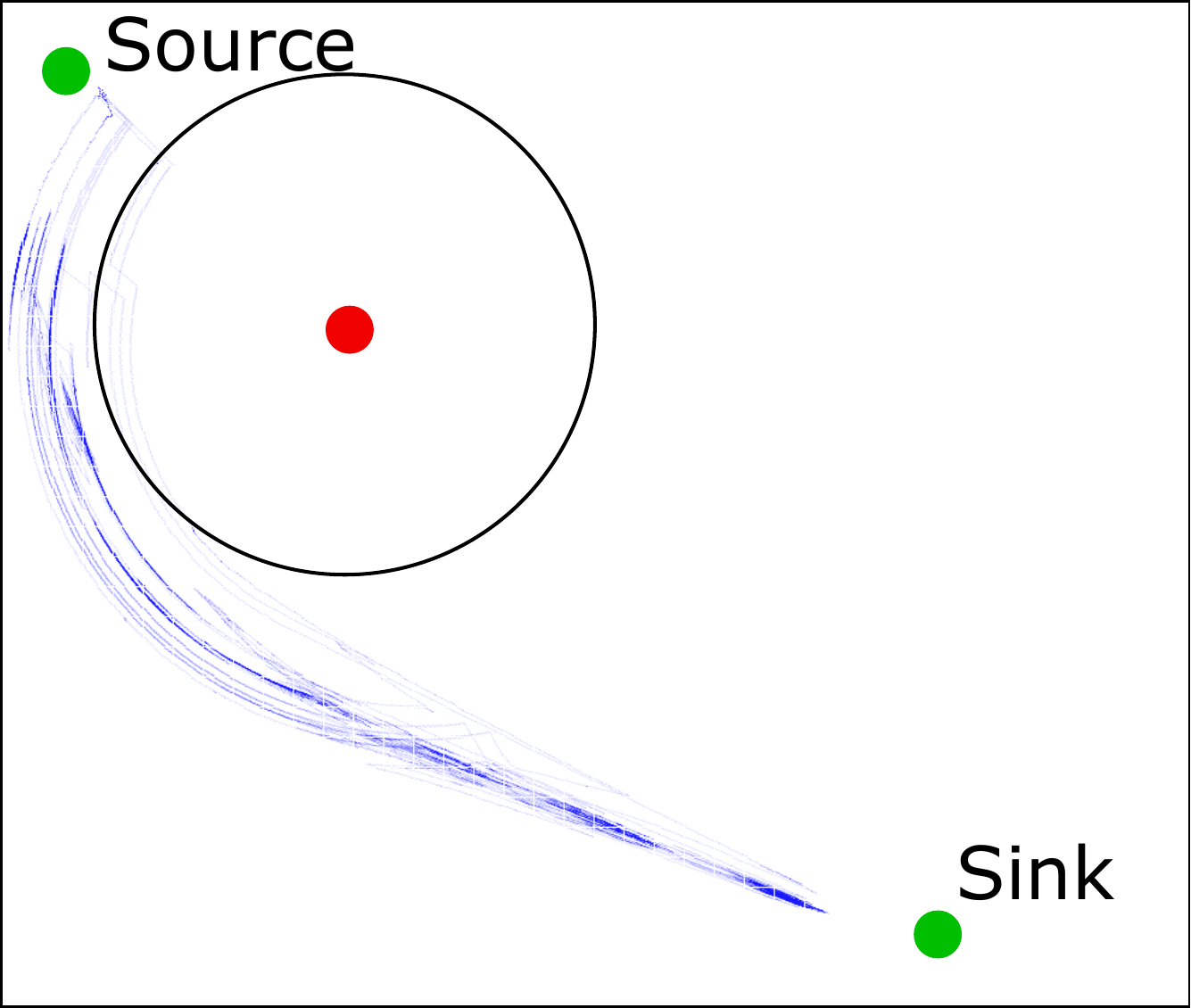} }
	\hspace*{2mm}
	\subfloat[]{    \includegraphics[width=0.25\linewidth]{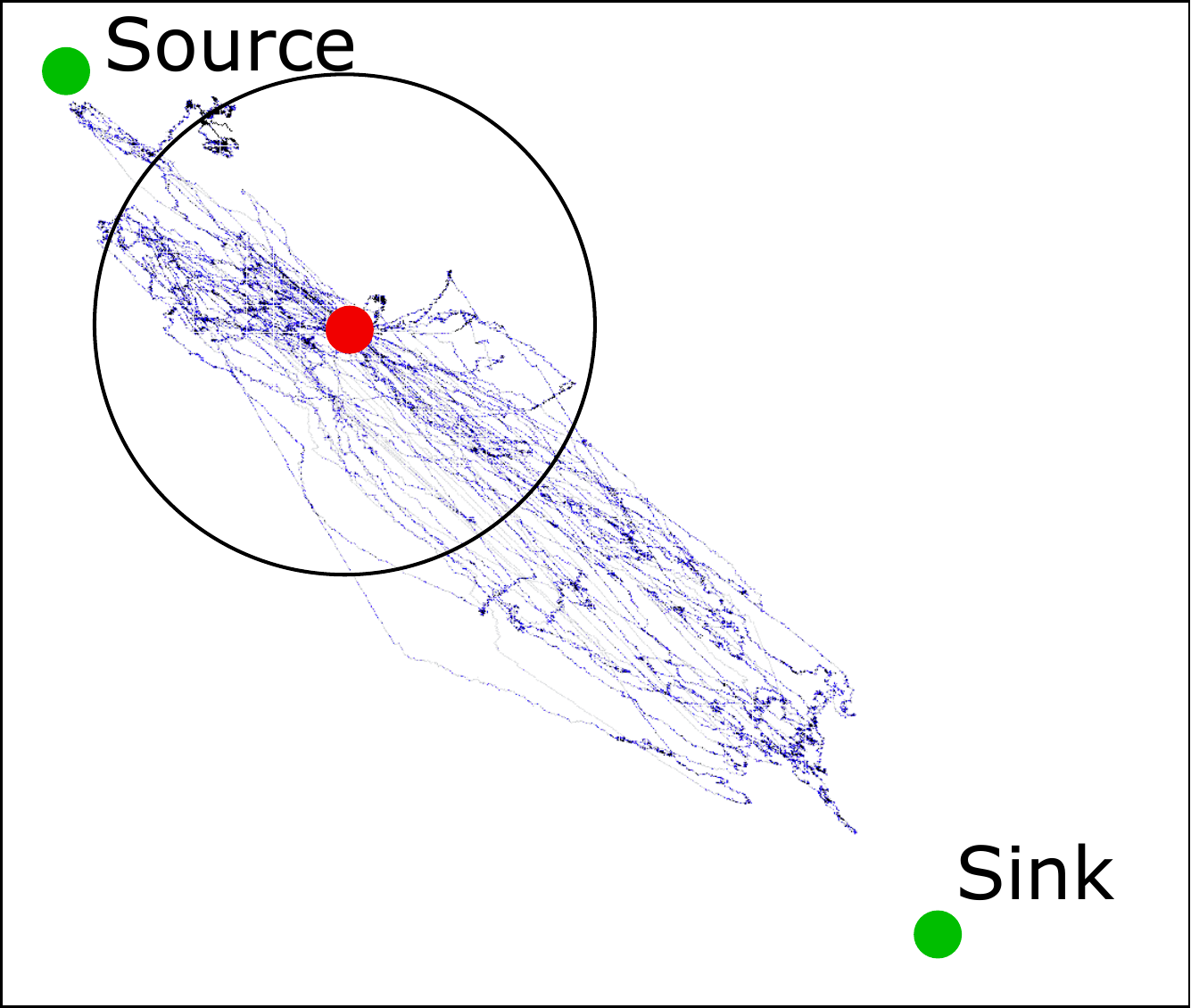} }\vspace*{-2mm}
	\caption{Motion tracking of swarm in a jammed, long-range environment. Source and Sink denoted by green circles, jammer with red circle, jammer estimated range as black ring, path of agents depicted with blue, semi-transparent lines. a) Designed behaviour uses repulsive forces to avoid jammer.  b) \gls{slcs2} swarm  exploring possible reactions to jamming device. Most of the swarm has become trapped by jammer restricting observability.} \label{jammeravoid}
	
\end{figure}

In regard to the jammed environments, the low impact of jamming devices on the designed behaviour is due to the swarm aggressively avoiding the jammer. This strong aversion, demonstrated in \Cref{jammeravoid} a), has been enforced by the designer as the negative impact of jammers was known \textit{a priori.} In contrast, \gls{slcs2} swarms do not create such strongly opposed behaviours, attempting to ferry over the jammer or brute-force communicate regardless of the noise, as seen in \Cref{jammeravoid} b). 
This low-quality evolution is due to the agents occasionally being trapped in the jammed area. Without the pre-designed repulsion, agents move into the jammed area as they travel toward the sink. The jammer interferes with neighbourhood discovery, which in turn leads to a restricted $\tilde{S}$ and thus limited $\Gamma''$. As such, the trapped agent cannot fully utilise the held rule-set to escape the jammer, nor can it correctly evaluate the held rules. This latter issue leads to incorrect rule assessment during evolution. Thus, jammers are seen to interfere with not only the agent operation but the evolution process as well.
%and thus limiting the rules observed to be valid via ($c_{n} \implies \tilde{S}$) and thus available for selection, as shown in \Cref{fig:jammrestrition}. %Agents thus move relative assumed neighbour positions, which are within the jamming area.
%    Should the remaining valid rules not allow escape from the jamming area the agent remain in the area for the remainder of the operation, unable to participate in the data-transfer task and only able to evaluate a small subset of $\Gamma$. This, in turn, restricts the evolution as the true strength and accuracy ($\epsilon$) of all rules in $\Gamma$ are not fully assessed.   


For the higher performance of \gls{slcs2} swarms over designed swarms in urban environments, inspection indicates this to be due to interconnected obstacle structures causing one or more human-designed swarm members to become trapped in concavities. This, in turn, leads to the entire swarm movement being halted, as fixed virtual force equations have the non-trapped agents continue to move relative to the trapped neighbours. In such situations, the swarm fails to reach the sink and a fitness of -1  is reported. Such failure was not observed during development of the behaviour, however, in 49.2\% of the 120 urban (and urban jammed) experiments such failure is reported. This demonstrates a significant limitation of human design, not being able to account for all environment eventualities.     This issue is not seen in the  \gls{slcs2} evolved behaviours, as the agents evolve mechanisms to avoid obstacle entrapment when required or agents learn to disregard trapped neighbours.

To conclude this main experiment, it has been shown that \gls{slcs2} produces swarm behaviours with greater mean fitness than the designed behaviour in most environment settings. In short-range open environments, this is due to the heterogeneous swarm achieving altruistic behaviours and techniques not thought of by the behaviour designer. In urban environments the success by  \gls{slcs2} 
is due to the evolution allowing problem instance specific behaviours, a feature likely too time-consuming for human designers. The one identified issue with \gls{slcs2}  is the failure due to jammers. This failure was found to not only impact the active solution of the swarm, but also the evolution process itself. As such, further work is required to allow \gls{slcs2} to utilise these pre-emptive danger aversions without impacting the flexibility of the evolution.

\subsection{Grammar specialisation}
\label{grammar_res}

\begin{figure}
	\centering
	\includestandalone[mode=image,height=2.00in]{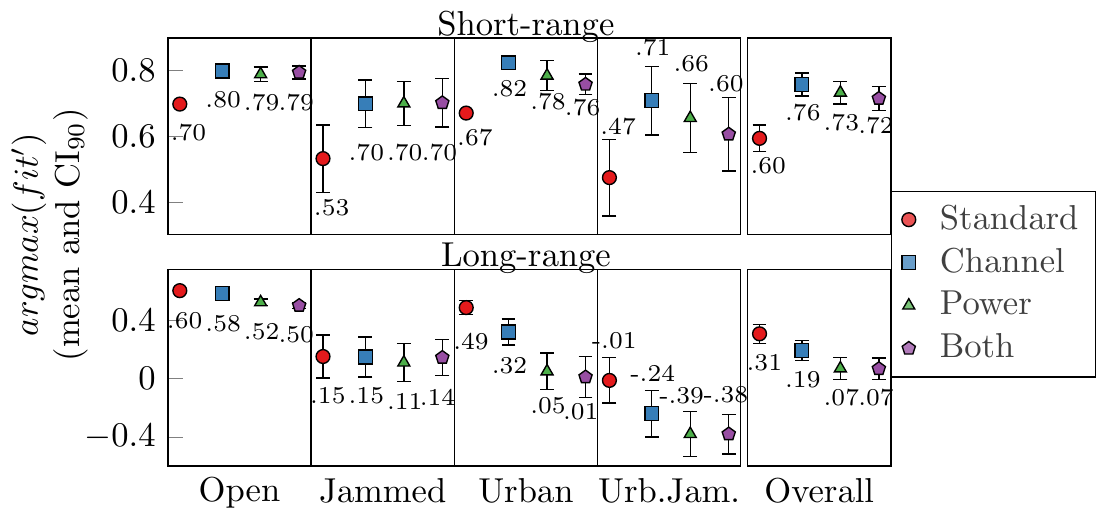}\vspace*{-2mm}
	\caption{Graphical results of mean fitness and \gls{ci} for standard and specialised grammars. Short-range top, long-range bottom. Each data point is the mean of 30 operations. Data labels rounded due to space restrictions. Alternative grammars are seen to produce higher mean fitnesses in short-range when congestion is being avoided, but long-range conditions have reduced fitness means.   }\label{g_results}
\end{figure}

For the third experiment of this study, \Cref{g_results} re-presents the result of \gls{slcs2} with the standard grammar and present \gls{slcs2} with the three specialised grammars. Appendinx E, again, shows the Mann-Whitney results. \Cref{g_results} show that in short-range environments all specialisations improve the evolved swarm performance with either significant or non-significant superiority (due to overlapping \gls{ci}). As such, mean-fitness improvements (compared to the standard grammar and relative to the full fitness range) of 8\%, 6.5\% and 6\% are respectively seen by the channel controlling, power controlling and combined alternative grammars. However, in long-range open and jammed environments, there is a negligible difference, and in the two long-range obstacle cases the standard approach achieves higher mean performance. Therefore, the overall mean fitness of the alternative grammars are lower than the standard grammar by, respectively, 6\%, 12\% and 12\% of the fitness range.  The remainder of this experiment investigation examines the lower performance by the specialised grammars in long-range cases and the improvement in short-range.

Examining the reduction in performance of specialised grammars in long-range environments, the power control grammar is first analysed. The behaviours of this grammar initially have $\sim$50\% of communication rules using the half-power setting. This ratio is expected from a two-option decoding (as shown in \Cref{TablePower}). It can be noted that this low-power setting is undesirable, as the required communication distance should have agents transmitting at maximum power to cover maximum range. Thus, it is expected such rules will be evolved out of the swarm. However, examining the communication rules at evolution end reveals this ratio remains (relatively) unchanged. This lack of alteration is found to be due to $q$, $\zeta$ and $\epsilon$ being similar for half- and full-power transmission actions, due to the stochastic network, intra-swarm interference and localised observations. Thus, the implemented evolution cannot remove the ineffective half-power rules. This failure observation is reinforced by the combined control grammar achieving relatively equal mean fitnesses in all eight environments.  This failure suggests tailoring the behaviour grammar for specific problems is either ineffective or would only be effective if significant alteration was made to the other evolution aspects, such as evaluation equations. 

In contrast to inappropriate action use, the reduction in performance by the channel grammar is due to the increase in behaviour-space size. Such growth in the behaviour-space results in exponential growth in the solution-space, and thus both the agent evolution and the solution novelty searches explore smaller percentages of the respective spaces in the 500 generations. As such, sub-optimal solution discoveries are made during \gls{slcs2} evolution with this grammar. To overcome this second failure, it is believed increasing the evolution generation limit may allow for greater discoveries at the cost of computation time.

Exploring the fitness improvements in short-range environments reveals the additional actions operate as intended; agents may communicate with limited congestion, without requiring non-optimal positioning (as seen in \Cref{short_range_comp}), and \gls{slcs2} achieves noticeably better performance around jammers.
However, examining all specialised grammars shows $q$, $\zeta$ and $\epsilon$ values for specialised and normal communication actions are, again, relatively similar.  It can, therefore, be concluded that the improvement of specialised grammars in short-range cases is not the result of evolution favouring these specialised rules, but rather the alternative actions being indistinguishable, and the agents' passive use leading to an emergent behaviour with greater fitness.

This analysis shows that the specialised grammars are effective in environments for which they were intended. However, in other environments, additional grammar features adversely affected swarm performance.
In essence, this failure of the evolution to utilise the specialised rules is due to the difference between the standard and specialised rules being less than the stochasticity of the environment. As such, the effects of the specialisation were passively included in the swarm and therefore the inclusion or exclusion of such features remained the responsibility of the designer, which is counter-intuitive to the autonomous approach of \gls{slcs2}.

\section{Conclusion}\label{conc}

The swarm behaviours autonomously evolved via \gls{slcs2} in this study are seen to effectively complete a data-transfer task in challenging conditions which include obstacles and jamming devices. To our knowledge, this problem domain is the most challenging seen in swarm evolution literature, and yet the proposed Lamarckian-style evolution of non-primitive rules, coupled with \gls{rl}, rule exchange between the swarm agents and novel search of the heterogeneous swarm behaviours has allowed the swarm to successfully complete the task. The former of these additions, rule exchange, saw a fitness improvement of up to 17\% and latter allowed up to 43\% improvement.

We compared \gls{slcs2} to a swarm behaviour designed by a human expert with a thorough understanding of the problem domain and swarm agents. Such a comparison was required as no other autonomous swarm behaviour algorithms are known to solve such a complex task.

The results of this comparison show that \gls{slcs2} swarms were able to out-perform the human-designed swarm with statistical significance in five of the eight environment settings, while the other three settings result in statistical ties. Furthermore, in the settings which \gls{slcs2} outperformed the designed swarm, a fitness range improvement of up to 31\% was observed.  From these results, it is concluded that \gls{slcs2} has bridged the gap between human-design and evolution. In doing so it has achieved the environmental flexibility of behaviour evolution and the high performance of human design.

In relation to further bridging this design-evolved behaviour gap, the behaviour creation grammar of \gls{slcs2} was specialised for the particular problem instances of this study. This specialisation resulted in a small, incidental improvement in performance for the intended environment setting, up to 8\%, but a greater reduction, of up to 12\%, was seen in the other environments. Hence, it is concluded that expanding the behaviour-space with problem-specific controls results in biased swarm behaviours and overly large behaviour-spaces, which could not be thoroughly explored without relaxing the evolution limitations (evolving for more generations). These failures resulted in reduced environment flexibility in the swarm. 

Finally, \gls{slcs2} experienced reduced evolution performance in some environments with jamming devices. In contrast,  the designed behaviour was able to utilise the intuition of the human designer and succeed via a strong aversive behaviour. As such, future iterations of the \gls{slcs} architecture aims to combine the insightful knowledge of designers with the flexibility, creativity and environment optimisation of \gls{slcs2}.

\section*{Acknowledgement}

Funding for this research was provided by Cyber and Electronic Warfare Division, Defence Science and Technology Group, Commonwealth of Australia.

\bibliography{thebib}    

\begin{thebibliography}{}

\bibitem[Beni, 2004]{beni2004swarm}
Beni, G. (2004).
\newblock From swarm intelligence to swarm robotics.
\newblock In {\em International Workshop on Swarm Robotics}, pages 1--9.
  Springer.

\bibitem[Castillo et~al., 2006]{castillo2006lamarckian}
Castillo, P., Arenas, M., Castellano, J., Merelo, J., Prieto, A., Rivas, V.,
  and Romero, G. (2006).
\newblock Lamarckian evolution and the baldwin effect in evolutionary neural
  networks.
\newblock {\em arXiv preprint cs/0603004}.

\bibitem[Duarte et~al., 2018]{duarte2018evolution}
Duarte, M., Gomes, J., Oliveira, S.~M., and Christensen, A.~L. (2018).
\newblock Evolution of repertoire-based control for robots with complex
  locomotor systems.
\newblock {\em IEEE Transactions on Evolutionary Computation}, 22(2):314--328.

\bibitem[Francesca et~al., 2015]{francesca2015}
Francesca, G., Brambilla, M., Brutschy, A., Garattoni, L., Miletitch, R.,
  Podevijn, G., Reina, A., Soleymani, T., Salvaro, M., Pinciroli, C., et~al.
  (2015).
\newblock Automode-chocolate: automatic design of control software for robot
  swarms.
\newblock {\em Swarm Intelligence}, 9(2-3):125--152.

\bibitem[Fraser et~al., 2017]{fraser2017simulating}
Fraser, B., Hunjet, R., and Szabo, C. (2017).
\newblock Simulating the effect of degraded wireless communications on emergent
  behavior.
\newblock In {\em 2017 Winter Simulation Conference (WSC)}, pages 4081--4092.
  IEEE.

\bibitem[Ghafoor et~al., 2014]{Ghafoor2014}
Ghafoor, K.~Z., Lloret, J., Sadiq, A.~S., and Mohammed, M.~A. (2014).
\newblock Improved geographical routing in vehicular ad hoc networks.
\newblock {\em Wireless Personal Communications}, 80(2):785--804.

\bibitem[Gomes et~al., 2015]{gomes2015cooperative}
Gomes, J., Mariano, P., and Christensen, A.~L. (2015).
\newblock Cooperative coevolution of partially heterogeneous multiagent
  systems.
\newblock In {\em Proceedings of the 2015 International Conference on
  Autonomous Agents and Multiagent Systems}, pages 297--305. International
  Foundation for Autonomous Agents and Multiagent Systems.

\bibitem[Gomes et~al., 2013]{gomes2013evolution}
Gomes, J., Urbano, P., and Christensen, A.~L. (2013).
\newblock Evolution of swarm robotics systems with novelty search.
\newblock {\em Swarm Intelligence}, 7(2-3):115--144.

\bibitem[Gordon and Grefenstette, 1995]{Gordon1995}
Gordon, D. and Grefenstette, J.~J. (1995).
\newblock Explanations of empirically derived reactive plans.
\newblock In {\em Proc. Seventh Int. Conf. on Machine Learning}, pages
  198--203.

\bibitem[Grefenstette, 1988]{grefenstette1988}
Grefenstette, J.~J. (1988).
\newblock Credit assignment in rule discovery systems based on genetic
  algorithms.
\newblock {\em Machine Learning}, 3(2-3):225--245.

\bibitem[Grefenstette, 1995]{grefenstette1995}
Grefenstette, J.~J. (1995).
\newblock Lamarckian learning in multi-agent environments.
\newblock Technical report, Navy Center for Applied Research in Artificial
  Intelligence Washinton DC.

\bibitem[Hauert et~al., 2009]{hauert2009}
Hauert, S., Zufferey, J.-C., and Floreano, D. (2009).
\newblock Evolved swarming without positioning information: an application in
  aerial communication relay.
\newblock {\em Autonomous Robots}, 26(1):21--32.

\bibitem[Heinerman et~al., 2015]{Heinerman2015b}
Heinerman, J., Drupsteen, D., and Eiben, A.~E. (2015).
\newblock Three-fold adaptivity in groups of robots: The effect of social
  learning.
\newblock In {\em Proceedings of the 2015 Annual Conference on Genetic and
  Evolutionary Computation}, pages 177--183. ACM.

\bibitem[Holland and Reitman, 1978]{Holland1978}
Holland, J.~H. and Reitman, J.~S. (1978).
\newblock Cognitive systems based on adaptive algorithms.
\newblock In {\em Pattern-directed inference systems}, pages 313--329.
  Elsevier.

\bibitem[Hurst et~al., 2002]{Hurst2002}
Hurst, J., Bull, L., and Melhuish, C. (2002).
\newblock Tcs learning classifier system controller on a real robot.
\newblock {\em Parallel problem solving from nature-PPSN VII}, pages 588--597.

\bibitem[Kelly and Heywood, 2018]{kelly2018emergent}
Kelly, S. and Heywood, M.~I. (2018).
\newblock Emergent solutions to high-dimensional multitask reinforcement
  learning.
\newblock {\em Evolutionary computation}, 26(3):347--380.

\bibitem[Luke et~al., 2005]{mason}
Luke, S., Cioffi-Revilla, C., Panait, L., Sullivan, K., and Balan, G. (2005).
\newblock Mason: A multiagent simulation environment.
\newblock {\em Simulation}, 81(7):517--527.

\bibitem[Medal, 2016]{medal2016}
Medal, H.~R. (2016).
\newblock The wireless network jamming problem subject to protocol
  interference.
\newblock {\em Networks}, 67(2):111--125.

\bibitem[Meyerson and Miikkulainen, 2017]{meyerson2017discovering}
Meyerson, E. and Miikkulainen, R. (2017).
\newblock Discovering evolutionary stepping stones through behavior domination.
\newblock In {\em Proceedings of the Genetic and Evolutionary Computation
  Conference}, pages 139--146. ACM.

\bibitem[Misir et~al., 2010]{Miser2010}
Misir, M., Verbeeck, K., De~Causmaecker, P., and Berghe, G.~V. (2010).
\newblock Hyper-heuristics with a dynamic heuristic set for the home care
  scheduling problem.
\newblock In {\em Evolutionary Computation (CEC), 2010 IEEE Congress on}, pages
  1--8.

\bibitem[Nelson et~al., 2004]{nelson2004}
Nelson, A.~L., Grant, E., and Henderson, T.~C. (2004).
\newblock Evolution of neural controllers for competitive game playing with
  teams of mobile robots.
\newblock {\em Robotics and Autonomous Systems}, 46(3):135--150.

\bibitem[O'Neill, 2003]{ONeill2003}
O'Neill, M. (2003).
\newblock {\em Grammatical evolution : evolutionary automatic programming in an
  arbitrary language}.
\newblock Boston : Kluwer Academic Publishers.

\bibitem[Perez-Uribe et~al., 2003]{perez2003}
Perez-Uribe, A., Floreano, D., and Keller, L. (2003).
\newblock Effects of group composition and level of selection in the evolution
  of cooperation in artificial ants.
\newblock In {\em European Conference on Artificial Life}, pages 128--137.
  Springer.

\bibitem[Ramanathan and Rosales-Hain, 2000]{ramanathan2000topology}
Ramanathan, R. and Rosales-Hain, R. (2000).
\newblock Topology control of multihop wireless networks using transmit power
  adjustment.
\newblock In {\em INFOCOM 2000. Nineteenth Annual Joint Conference of the IEEE
  Computer and Communications Societies. Proceedings. IEEE}, volume~2, pages
  404--413. IEEE.

\bibitem[Resende and Ribeiro, 2016]{resende2016optimization}
Resende, M.~G. and Ribeiro, C.~C. (2016).
\newblock {\em Optimization by GRASP: Greedy Randomized Adaptive Search
  Procedures}.
\newblock Springer.

\bibitem[Silva et~al., 2016]{silva2016open}
Silva, F., Duarte, M., Correia, L., Oliveira, S.~M., and Christensen, A.~L.
  (2016).
\newblock Open issues in evolutionary robotics.
\newblock {\em Evolutionary computation}, 24(2):205--236.

\bibitem[Smith and Congdon, 2005]{smith2005rcs}
Smith, N.~W. and Congdon, C.~B. (2005).
\newblock Rcs: a learning classifier system for evolutionary robotics.
\newblock In {\em GECCO Workshops}, pages 119--120. Citeseer.

\bibitem[Smith et~al., 2018]{smith2018data}
Smith, P., Hunjet, R., Aleti, A., Barca, J.~C., et~al. (2018).
\newblock Data transfer via uav swarm behaviours: Rule generation, evolution
  and learning.
\newblock {\em Australian Journal of Telecommunications and the Digital
  Economy}, 6(2):35--58.

\bibitem[Smith, 1980]{smith1980learning}
Smith, S.~F. (1980).
\newblock {\em A learning system based on genetic adaptive algorithms}.
\newblock PhD thesis, University of Pittsburgh.

\bibitem[Soleymani et~al., 2015]{Soleymani2015}
Soleymani, T., Trianni, V., Bonani, M., Mondada, F., and Dorigo, M. (2015).
\newblock Bio-inspired construction with mobile robots and compliant pockets.
\newblock {\em Robotics and Autonomous Systems}, 74:340--350.

\bibitem[Szepesv{\'a}ri, 2010]{Szepesvari2010}
Szepesv{\'a}ri, C. (2010).
\newblock Algorithms for reinforcement learning.
\newblock {\em Synthesis lectures on artificial intelligence and machine
  learning}, 4(1):1--103.

\bibitem[Takadama et~al., 2000]{takadama2000}
Takadama, K., Terano, T., and Shimohara, K. (2000).
\newblock Learning classifier systems meet multiagent environments.
\newblock In {\em International Workshop on Learning Classifier Systems}, pages
  192--210. Springer.

\bibitem[Timmis et~al., 2016]{Timmis2016}
Timmis, J., Ismail, A.~R., Bjerknes, J.~D., and Winfield, A.~F. (2016).
\newblock An immune-inspired swarm aggregation algorithm for self-healing swarm
  robotic systems.
\newblock {\em Biosystems}, 146:60--76.

\bibitem[Trianni, 2008]{trianni2008evolutionary}
Trianni, V. (2008).
\newblock {\em Evolutionary swarm robotics: evolving self-organising behaviours
  in groups of autonomous robots}, volume 108.
\newblock Springer.

\bibitem[Watkins, 1989]{watkins1989learning}
Watkins, C. J. C.~H. (1989).
\newblock {\em Learning from delayed rewards}.
\newblock PhD thesis, King's College, Cambridge.

\bibitem[Wilson, 1978]{wilson1978division}
Wilson, E.~O. (1978).
\newblock Division of labor in fire ants based on physical castes (hymenoptera:
  Formicidae: Solenopsis).
\newblock {\em Journal of the Kansas Entomological Society}, pages 615--636.

\bibitem[Wilson, 1995]{Wilson1995}
Wilson, S.~W. (1995).
\newblock Classifier fitness based on accuracy.
\newblock {\em Evolutionary computation}, 3(2):149--175.

\end{thebibliography}

\end{document}